\documentclass{article}

     \usepackage[preprint]{neurips_2020}

\usepackage[utf8]{inputenc} %
\usepackage[T1]{fontenc}    %
\usepackage{url}            %
\usepackage{booktabs}       %
\usepackage{amsfonts}       %
\usepackage{nicefrac}       %
\usepackage{microtype}      %

\usepackage{amsmath}
\usepackage{algorithm,algorithmic} 
\usepackage{microtype}
\usepackage[hang,flushmargin]{footmisc} 

\usepackage{color}
\usepackage{graphicx}
\graphicspath{{figures/}}
\usepackage{subfigure}
\usepackage{wrapfig}
\usepackage{booktabs} %
\usepackage{multirow,multicol}
\definecolor{mydarkblue}{rgb}{0,0.08,0.45}
\usepackage[colorlinks,citecolor=mydarkblue,urlcolor=mydarkblue,linkcolor=mydarkblue,filecolor=mydarkblue]{hyperref}

\usepackage{amsmath,amsfonts,bm}

\def\eqref#1{equation~\ref{#1}}

\def\1{\bm{1}}

\def\vmu{{\bm{\mu}}}
\def\vtheta{{\bm{\theta}}}

\def\vv{{\bm{v}}}

\def\vx{{\bm{x}}}

\DeclareMathAlphabet{\mathsfit}{\encodingdefault}{\sfdefault}{m}{sl}
\SetMathAlphabet{\mathsfit}{bold}{\encodingdefault}{\sfdefault}{bx}{n}

\newcommand{\KL}{D_{\mathrm{KL}}}

\DeclareMathOperator*{\argmin}{arg\,min}

\newcommand{\Reals}{\mathbb{R}}

\newcommand{\data}{\{\vx_n, y_n\}_{n=1}^N}
\newcommand{\map}{\mathsf{MAP}}

\newcommand{\email}[1]{\footnotesize{\href{mailto:#1}{\url{#1}} }} 

\usepackage{enumitem}
\setlist[enumerate]{leftmargin=10pt} %
\setlist[itemize]{leftmargin=10pt} %

 \newlength{\sfigwidth}
 \newlength{\sfigwidthtwo}
 \newlength{\sfigwidththree}
 \newlength{\sfigwidththreesubfigure}
 \newlength{\sfigwidthtsne}
\newlength{\sfigwidthimagenet}
 \newlength{\sfigwidthlandscape}
 \usepackage[compact]{titlesec}
 \titlespacing{\section}{0pt}{1ex}{0ex}
 \titlespacing{\subsection}{0pt}{1ex}{0ex}
 \titlespacing{\subsubsection}{0pt}{0.5ex}{0ex}
 \newcommand{\reducespaceafterfigure}{\vspace{-1em}} %
 \newcommand{\reducevspace}{\vspace{-1em}} %

\title{Deep Ensembles: A Loss Landscape Perspective}

\author{Stanislav Fort\thanks{Equal contribution.}
\\
Google Research\\
\email{sfort1@stanford.edu} \\
\And
Huiyi Hu\footnotemark[1] \\
DeepMind \\
\email{clarahu@google.com}
\And
Balaji Lakshminarayanan\thanks{Corresponding author.  %
} \\
DeepMind \\
\email{balajiln@google.com}
}

\begin{document}

\maketitle

 \begin{abstract}
Deep ensembles have been empirically shown to be a promising approach for improving accuracy, uncertainty  and out-of-distribution robustness of deep learning models. While deep ensembles were theoretically motivated by the bootstrap, non-bootstrap ensembles trained with just random initialization also perform well in practice, which suggests that there could be other explanations for why deep ensembles work well. Bayesian neural networks, which learn distributions over the parameters of the network, are theoretically well-motivated by Bayesian principles, but do not perform as well as deep ensembles in practice, particularly under dataset shift. One possible explanation for this gap between theory and practice is that popular scalable variational Bayesian methods tend to focus on a single mode, whereas deep ensembles tend to explore diverse modes in function space. We investigate this hypothesis by building on recent work on understanding the loss landscape of neural networks and adding our own exploration to measure the similarity of functions in the space of predictions. Our results show that random initializations explore entirely different modes, while functions along an optimization trajectory or sampled from the subspace thereof cluster within a single mode predictions-wise, while often deviating significantly in the weight space. 
Developing the concept of the diversity--accuracy plane, we show that the decorrelation power of random initializations is unmatched by popular subspace sampling methods. Finally, we evaluate the relative effects of ensembling, subspace based methods and ensembles of subspace based methods, %
and the experimental results validate our hypothesis.  %
 \end{abstract}

\section{Introduction}

Consider a typical classification problem, where  $\vx_n\in\Reals^D$ denotes the $D$-dimensional features and $y_n\in[1,\ldots,K]$ denotes the class label. 
Assume we have a parametric model $p(y| \vx, \vtheta)$ for the conditional distribution where $\vtheta$ denotes weights and biases of a neural network, and $p(\vtheta)$ is a prior distribution over parameters. 
The Bayesian posterior over parameters is given by
$p(\vtheta| \{\vx_n,y_n\}_{n=1}^N) \propto p(\vtheta) \prod_{n=1}^N p(y_n | \vx_n, \vtheta)$.

Computing the exact posterior distribution over $\vtheta$ is  computationally expensive (if not impossible) when $p(y_n | \vx_n, \vtheta)$ is a deep neural network (NN). A variety of approximations have been developed for \emph{Bayesian neural networks}, including Laplace approximation \citep{mackay1992bayesian},  Markov chain Monte Carlo methods \citep{Neal96,Welling2011,springenberg2016bayesian}, variational Bayesian methods \citep{graves,BBB,louizos2017multiplicative,flipout} and Monte-Carlo dropout \citep{mcdropout,dropout}.  %
While computing the posterior is challenging, it is usually easy to perform maximum-a-posteriori (MAP) estimation, which corresponds to a mode of the posterior. The MAP solution can be written as the minimizer of the following loss: 
\vspace{-1em} 
\begin{align}
     \hat{\vtheta}_\map &= \argmin_\vtheta L(\vtheta, \data) %
    = \argmin_\vtheta
    -\log p(\vtheta) - \sum_{n=1}^N \log p(y_n | \vx_n, \vtheta). \label{eqn:loss} 
\end{align}
The MAP solution is computationally efficient, but only gives a point estimate and not a distribution over parameters. 
\emph{Deep ensembles}, proposed by \citet{lakshminarayanan2017simple}, train an ensemble of neural networks by initializing at $M$ different values and repeating the minimization multiple times %
which could lead to $M$ different solutions, if the loss is non-convex. 
\citet{lakshminarayanan2017simple} found adversarial training provides additional benefits in some of their experiments, but we will ignore adversarial training  and focus only on ensembles with random initialization. %

Given finite training data, many parameter values could equally well explain the observations, and capturing these diverse solutions is crucial for quantifying \emph{epistemic uncertainty} \citep{kendall2017uncertainties}. 
Bayesian neural networks learn a distribution over weights, and a good posterior approximation should be able to learn multi-modal posterior distributions in theory.  
Deep ensembles were inspired by the bootstrap \citep{bagging}, which has useful theoretical properties. However, it has been empirically observed by \citet{lakshminarayanan2017simple,mheads} that training individual networks with just random initialization is sufficient in practice and using the bootstrap can even hurt  performance (e.g. for small ensemble sizes). 
Furthermore, 
 \citet{ovadia2019can} and \citet{gustafsson2019evaluating} independently benchmarked existing methods for uncertainty quantification on a variety of datasets and architectures, and observed that ensembles tend to outperform  approximate Bayesian neural networks in terms of both accuracy and uncertainty, particularly under dataset shift. 

\begin{wrapfigure}{r}{0.5\textwidth}
\vspace{-1em}
    \centering%
      \includegraphics[width=0.475\textwidth]{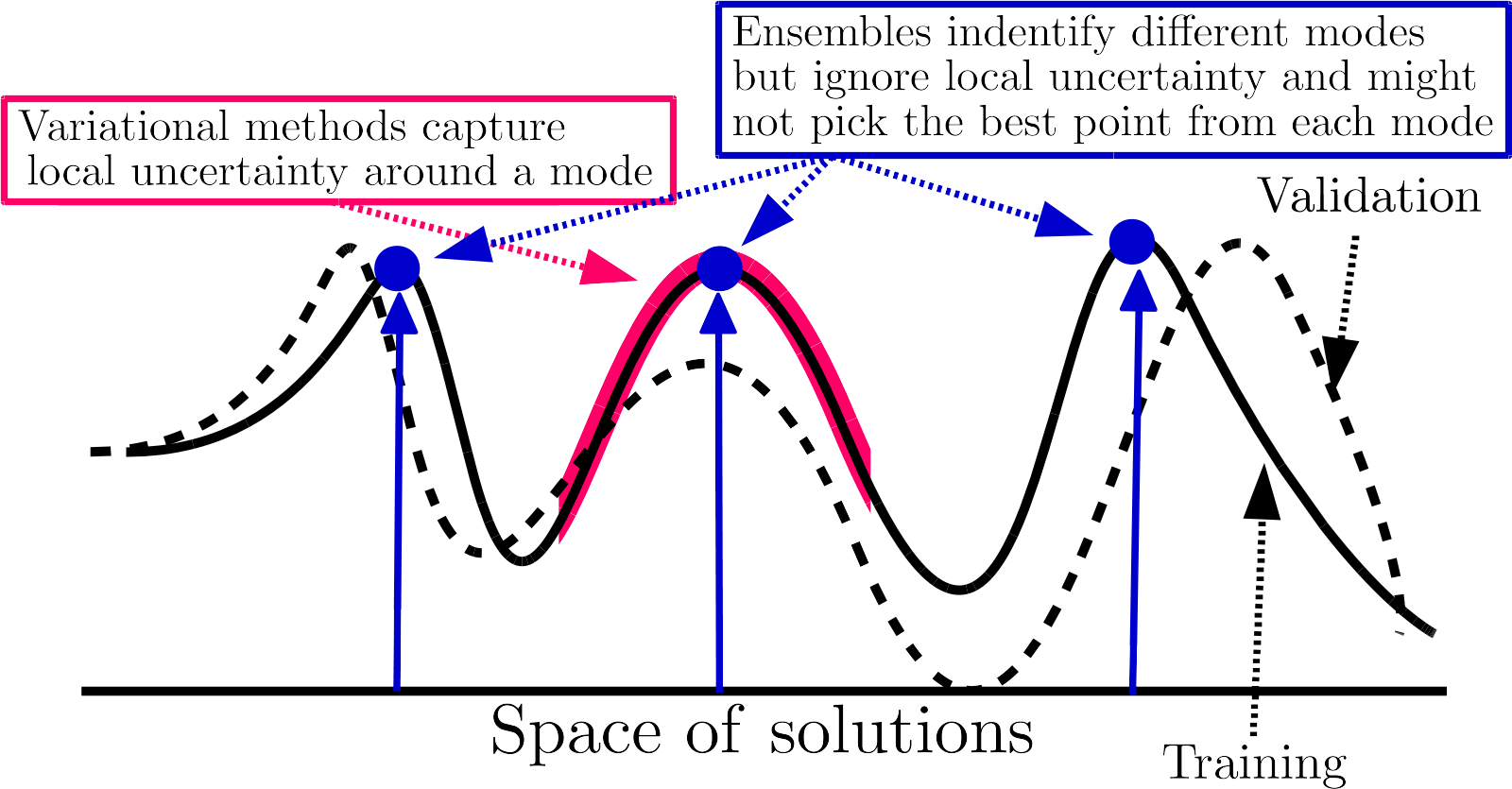}
\vspace{-0.5em}
    \caption{Cartoon illustration of the hypothesis. $x$-axis indicates parameter values and $y$-axis plots the negative loss $-L(\vtheta, \data)$ %
    on train and validation data. %
    }
    \vspace{-0.5em}
    \label{fig:ensemble-vs-bayes}%
\end{wrapfigure}
These empirical observations raise an important question:  %
   \emph{Why do %
deep ensembles trained with just random initialization 
work so well 
in practice?}  
One possible hypothesis is that ensembles tend to sample from different modes\footnote{We  use the term `mode' to refer to unique functions $f_{\vtheta}(\vx)$. Due to weight space symmetries, different parameters can correspond to the same function, i.e. $f_{\vtheta_1}(\vx) = f_{\vtheta_2}(\vx)$ even though $\vtheta_1\neq\vtheta_2$. %
} in function space, whereas variational Bayesian methods (which minimize $\KL(q(\vtheta)|p(\vtheta|\{\vx_n,y_n\}_{n=1}^N)$) might fail to explore multiple modes even though they %
are effective at capturing uncertainty within a single mode. 
See Figure~\ref{fig:ensemble-vs-bayes} for a cartoon illustration. Note that while the MAP solution is a local optimum for the training loss,%
it may not necessarily be a local optimum for the validation loss.

Recent work on understanding loss landscapes \citep{garipov2018loss,draxler2018essentially,fort2019large} allows us to investigate this hypothesis.  
Note that prior work on loss landscapes has focused on mode-connectivity and low-loss tunnels, but has not explicitly focused on how diverse the functions from different modes are. %
{The experiments in these papers (as well as other papers on deep ensembles) provide indirect evidence for this hypothesis, either through downstream metrics (e.g. accuracy and calibration) or by visualizing the performance along the low-loss tunnel.   
} We complement these works by \emph{explicitly measuring function space diversity within training trajectories and subspaces thereof (dropout, diagonal Gaussian, low-rank Gaussian and random  subspaces) and across different randomly initialized trajectories across multiple datasets, architectures, and dataset shift.}
Our findings show that the functions sampled along a single training trajectory or subspace thereof tend to be very similar in predictions (while potential far away in the weight space), whereas functions sampled from different randomly initialized trajectories tend to be very diverse. 

\section{Background}

The loss landscape of neural networks (also called the objective landscape) -- the space of weights and biases that the network navigates during training -- is a high dimensional function and therefore could potentially be very complicated. However, many empirical results show surprisingly simple properties of the loss surface. \citet{Goodfellow2014QualitativelyCN} observed that the loss along a linear path from an initialization to the corresponding optimum is monotonically decreasing, encountering no significant obstacles along the way. \citet{li2018measuring} demonstrated that constraining optimization to a random, low-dimensional hyperplane in the weight space leads to results comparable to full-space optimization, provided that the dimension exceeds a modest threshold. This was geometrically understood and extended by \citet{fort2019goldilocks}.

\citet{garipov2018loss,draxler2018essentially} demonstrate that while a linear path between two independent optima hits a high loss area in the middle, there in fact exist continuous, low-loss paths connecting any pair of optima (or at least any pair empirically studied so far). These observations are unified into a single phenomenological model in \citep{fort2019large}. %
While low-loss tunnels create functions with near-identical low values of loss along the path, the experiments of  \citet{fort2019large,garipov2018loss} provide preliminary evidence that these functions tend to be very different in function space, changing significantly in the middle of the tunnel, see Appendix~\ref{sec:diversity-along-tunnel} for a review and additional empirical evidence that complements their results.

\section{Experimental setup}

We explored the CIFAR-10 \citep{cifar10}, CIFAR-100 \citep{cifar10}, and ImageNet \citep{imagenet_cvpr09} datasets. We train convolutional neural networks on the CIFAR-10 dataset, which contains 50K training examples from 10 classes. To verify that our findings translate across architectures,  we use the following 3 architectures on CIFAR-10: %
\begin{itemize}\itemsep0em
    \item \emph{SmallCNN}: channels [16,32,32]
 for 10 epochs which achieves $64\%$ test accuracy.
     \item \emph{MediumCNN}: channels [64,128,256,256]
 for 40 epochs which achieves $71\%$ test accuracy.
    \item \emph{ResNet20v1} \citep{resnet}:
 for 200 epochs which achieves $90\%$ test accuracy.
\end{itemize}
We use the Adam optimizer \citep{adam} for training and to make sure the effects we observe are general, we validate that our results hold for vanilla stochastic gradient descent (SGD) as well (not shown due to space limitations). We use batch size 128 and dropout 0.1 for training  \emph{SmallCNN} and \emph{MediumCNN}. We used 40 epochs of training for each.
To generate weight space and prediction space similarity results, we use a constant learning rate of $1.6 \times 10^{-3}$ and halfing it every $10$ epochs, unless specified otherwise. We do not use any data augmentation with those two architectures. For \emph{ResNet20v1}, we use the data augmentation and learning rate schedule used in Keras examples.\footnote{\url{https://keras.io/examples/cifar10_resnet/}} The overall trends are consistent across all architectures, datasets, and other hyperparameter and non-linearity choices we explored.

To test if our observations generalize to other datasets, we also ran certain experiments on more complex datasets such as CIFAR-100 \citep{cifar10} which contains 50K examples belonging to 100 classes and ImageNet \citep{imagenet_cvpr09}, which contains roughly 1M examples belonging to 1000 classes. 
CIFAR-100 is trained using the same \emph{ResNet20v1} as above with Adam optimizer, batch size 128 and total epochs of 200. The learning rate starts from $10^{-3}$ and decays to $(10^{-4}, 5\times10^{-5},10^{-5},5\times10^{-7})$ at epochs $(100, 130, 160, 190)$. ImageNet is trained with \emph{ResNet50v2} %
\citep{he2016identity} and momentum optimizer ($0.9$ momentum), with batch size 256 and 160 epochs. The learning rate starts from $0.15$ and decays to $(0.015, 0.0015)$ at epochs $(80, 120)$.

 In addition to evaluating on regular test data, we also evaluate the performance of the methods on corrupted versions of the dataset using the CIFAR-10-C and ImageNet-C benchmarks  \citep{hendrycks2018benchmarking} which contain corrupted versions of original images with 19 corruption types (15 for ImageNet-C) and varying intensity values (1-5), and was used by \citet{ovadia2019can} to measure calibration of the uncertainty estimates under dataset shift. Following \citet{ovadia2019can}, we measure  accuracy as well as Brier score \citep{brier1950verification} (lower values indicate better uncertainty estimates). %
 {We use the SVHN dataset \citep{svhn} to evaluate how different methods trained on CIFAR-10 dataset react to out-of-distribution (OOD) inputs.}

\begin{figure*}[t]%
    \centering%
    \begin{subfigure}[Cosine similarity of weights %
    ]{
     \includegraphics[width=0.33\textwidth]{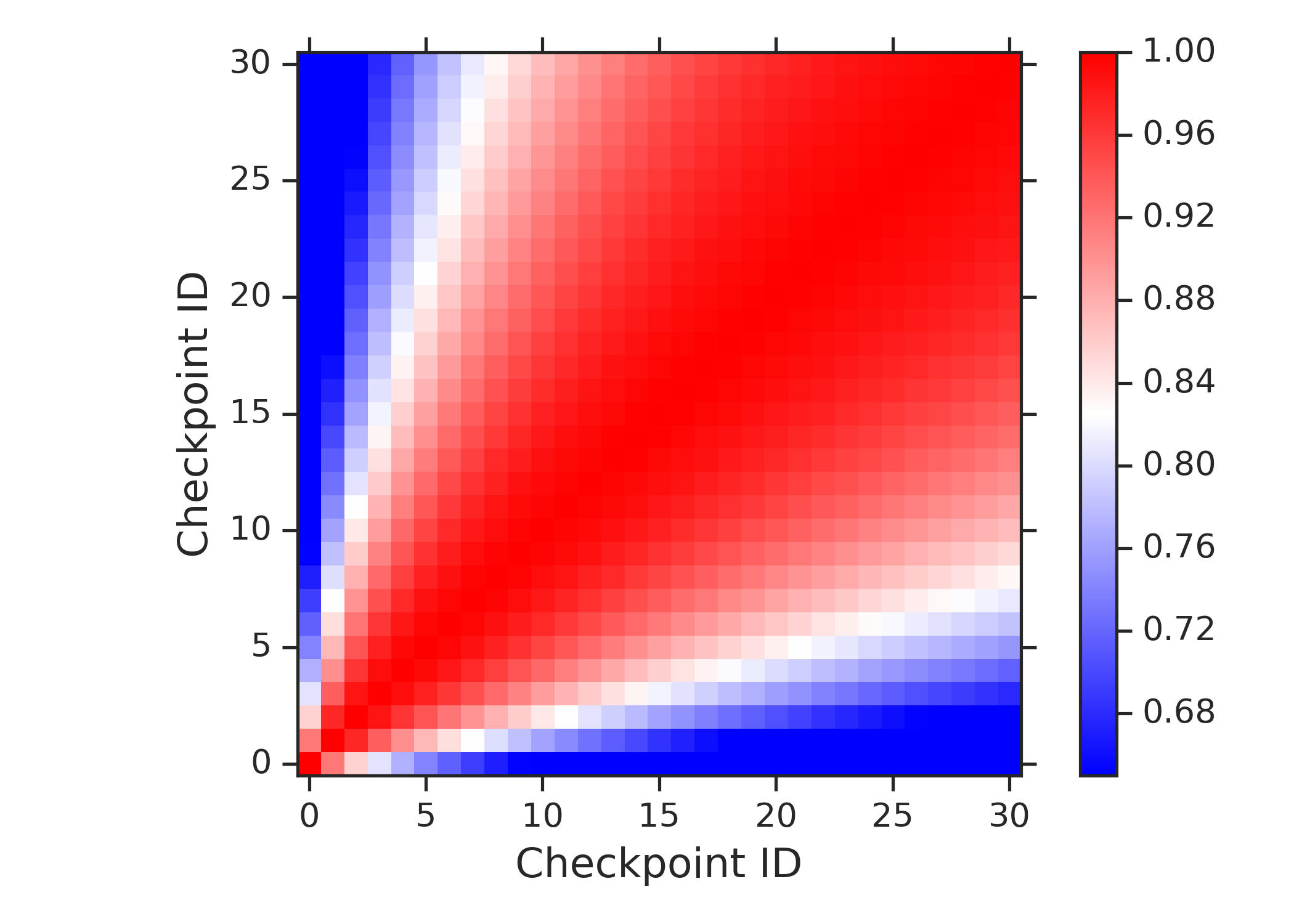}%
     \label{fig:cosine:smallcnn:withintrajectory}%
    }\end{subfigure} 
    \begin{subfigure}[Disagreement of predictions %
    ]{
     \includegraphics[width=0.33\textwidth]{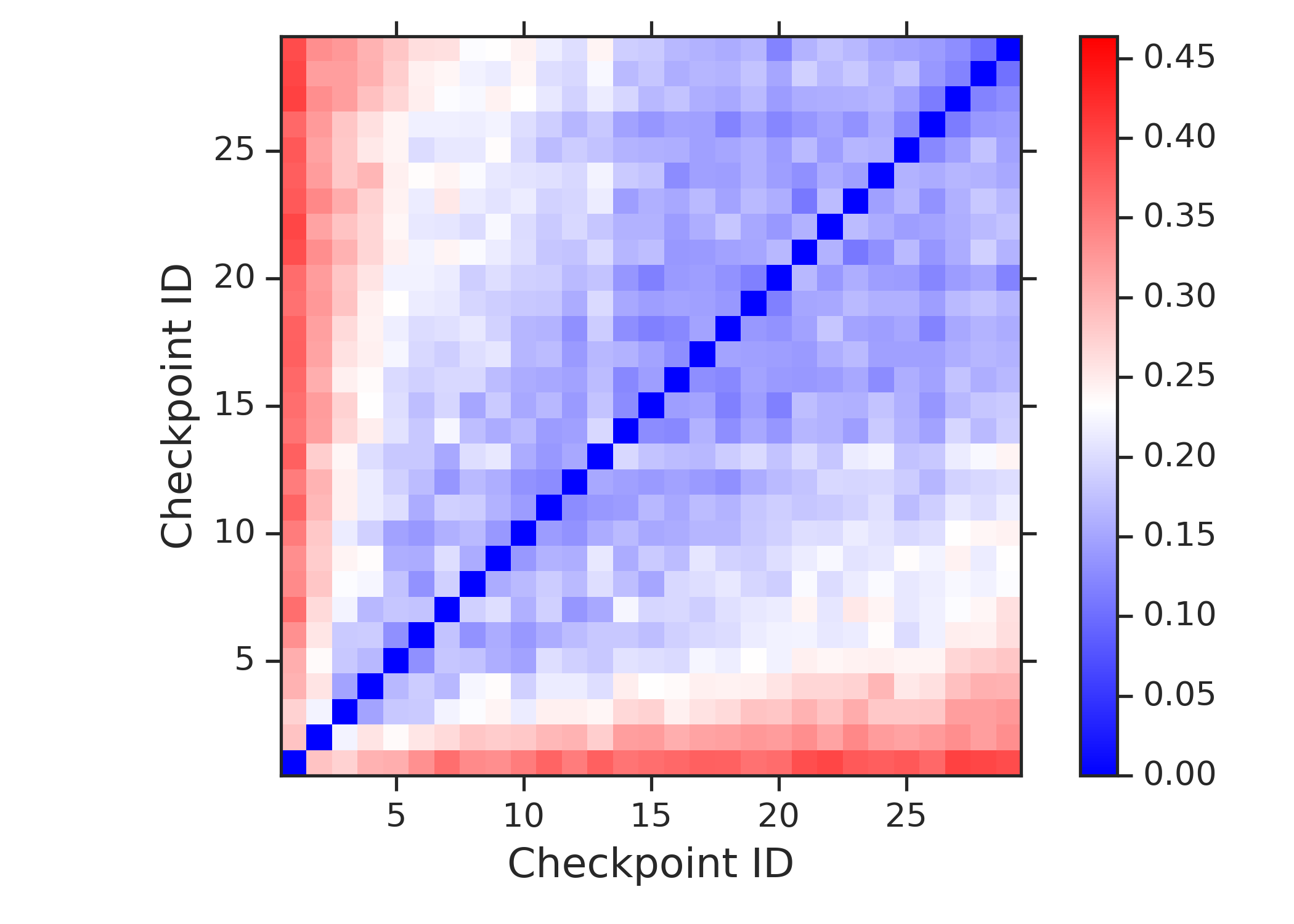}%
     \label{fig:fracdiff:smallcnn:withintrajectory}%
    }\end{subfigure} %
    \begin{subfigure}[t-SNE of predictions]{  
    \includegraphics[width=0.3\textwidth]{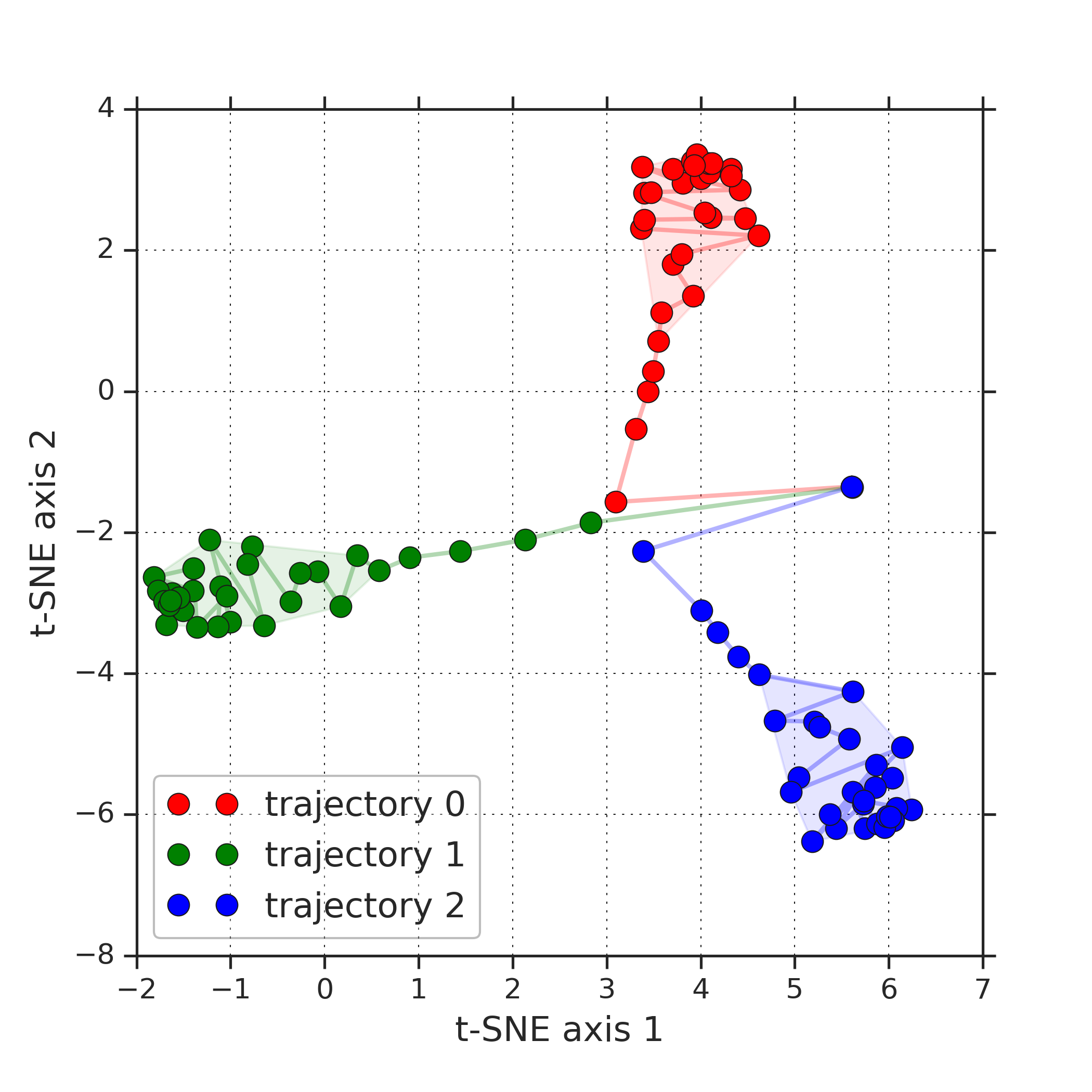}
        \label{fig:tsne-init}%
    }\end{subfigure}
   \reducespaceafterfigure
    \caption{Results using SmallCNN on CIFAR-10.
  \emph{Left plot}: Cosine similarity between checkpoints to measure weight space alignment along optimization trajectory.  \emph{Middle plot}: The fraction of labels on which the predictions from  different checkpoints disagree.  \emph{Right plot}: t-SNE plot of predictions from checkpoints corresponding to 3 different randomly initialized trajectories (in different colors). 
    }
    \label{fig:smallcnn-similarity-tsne}%
\end{figure*}%

 \section{Visualizing Function Space Similarity} 
\subsection{Similarity of Functions Within and Across Randomly Initialized Trajectories}

\begin{figure}[ht]
\begin{center}
   \begin{subfigure}[Results using \emph{SmallCNN}]{  \includegraphics[width=0.475\textwidth]{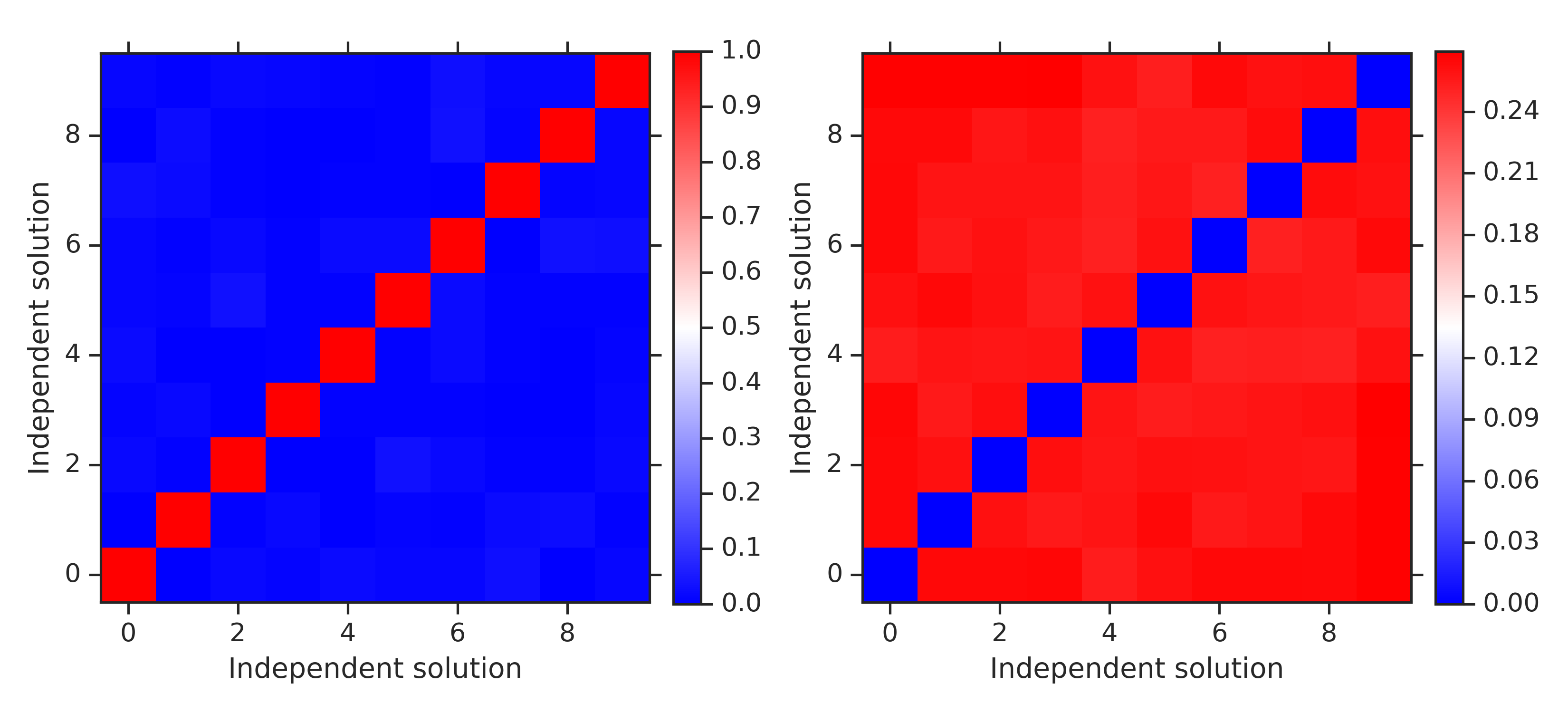}
   }\end{subfigure}
    \begin{subfigure}[Results using \emph{ResNet20v1}]{ \includegraphics[width=0.475\textwidth]{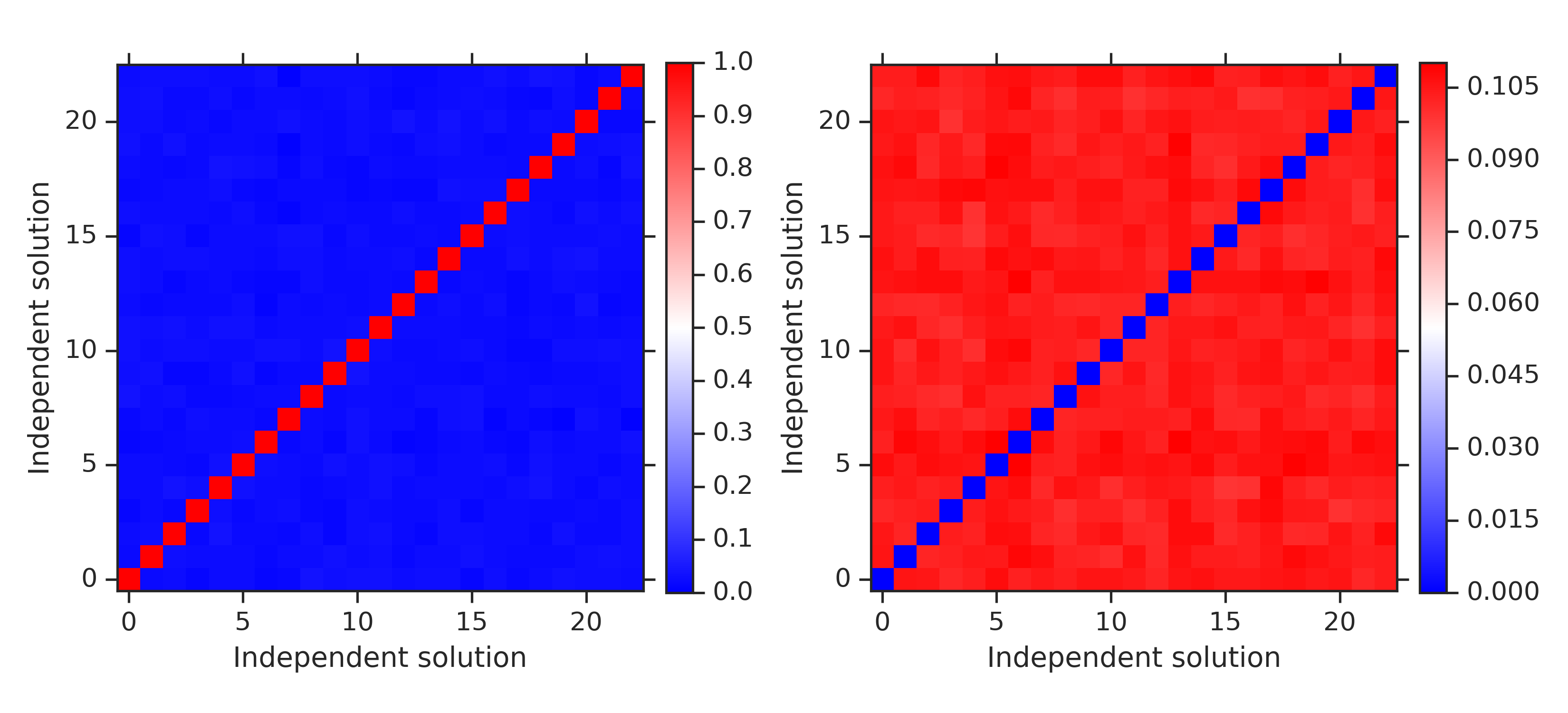}
    }\end{subfigure}
    \reducespaceafterfigure
      \caption{Results on CIFAR-10 using two different architectures.  For each of these architectures, the left subplot shows the  
   cosine similarity between different solutions in weight space, and the right subplot shows the 
      fraction of labels on which the predictions from  different solutions disagree. In general, the weight vectors from two different initializations are essentially orthogonal, while their predictions are approximately as dissimilar as any other pair. 
      }
     \label{fig:acrossinit}%
      \end{center}
\end{figure}

First, we compute the similarity between different checkpoints along a single trajectory. In Figure~\ref{fig:cosine:smallcnn:withintrajectory}, we plot the cosine similarity in weight space, defined as $cos(\vtheta_1,\vtheta_2)=\frac{\vtheta_1^\top\vtheta_2}{||\vtheta_1||||\vtheta_2||}$.  
In Figure~\ref{fig:fracdiff:smallcnn:withintrajectory}, we plot the disagreement in function space, defined as the fraction of points the checkpoints disagree on, that is, $\frac{1}{N}\sum_{n=1}^N [f(\vx_n;\vtheta_1)\neq f(\vx_n;\vtheta_2)]$, where $f(\vx;\vtheta)$ denotes the class label predicted by the network for input $\vx$. 
We observe that the checkpoints along a trajectory are largely similar both in the weight space and the function space.
Next, we evaluate how diverse the final solutions from different random initializations are. 
The functions from different initialization are different,  as demonstrated by the similarity plots in 
Figure~\ref{fig:acrossinit}. Comparing this with Figures~\ref{fig:cosine:smallcnn:withintrajectory} and \ref{fig:fracdiff:smallcnn:withintrajectory}, we see that functions within a single trajectory exhibit higher similarity  and functions across  different trajectories exhibit much lower similarity. %

Next, we take the predictions from different checkpoints along the individual training trajectories from multiple initializations and compute a t-SNE plot \citep{maaten2008visualizing} to visualize their similarity in function space. More precisely, for each checkpoint we take the softmax output for a set of examples, flatten the vector and use it to represent the model's predictions. The t-SNE algorithm is then used to reduce it to a 2D point in the t-SNE plot. 
Figure~\ref{fig:tsne-init} shows that the functions explored by different trajectories (denoted by circles with different colors) are far away, while functions explored within a single trajectory (circles with the same color) tend to be much more similar.

\reducevspace
\subsection{Similarity of Functions Within Subspace from Each Trajectory and Across Trajectories}
\label{sec:fun_similarity_subspace}
In addition to the checkpoints along a trajectory, we also construct subspaces based on each individual trajectory.  %
Scalable variational Bayesian methods typically approximate the distribution of weights along the training trajectory, hence visualizing the diversity of functions between the subspaces helps understand the difference between  Bayesian neural networks and ensembles.
We use a representative set of four subspace sampling methods: Monte Carlo dropout, a diagonal Gaussian approximation, a low-rank covariance matrix Gaussian approximation and a random subspace approximation.  
{Unlike dropout and Gaussian approximations which assume a parametric form for the variational posterior, the random subspace method explores all high-quality solutions within the subspace and hence could be thought of as a non-parametric variational approximation to the posterior.}  Due to space constraints, we do not consider Markov Chain Monte Carlo (MCMC) methods in this work; \citet{Zhang2020Cyclical} show that popular stochastic gradient MCMC (SGMCMC) methods may not explore multiple modes and propose cyclic SGMCMC. We compare diversity of random initialization and cyclic SGMCMC in Appendix~\ref{sec:csgmcmc}. 
In the descriptions of the methods, let $\vtheta_0$ be the optimized weight-space solution (the weights and biases of our trained neural net) around which we will construct the subspace. %
\begin{itemize}\itemsep0em
    \item \textbf{Random subspace sampling}: We start at an optimized solution $\vtheta_0$ and choose a random direction $\vv$ in the weight space. We step in that direction by choosing different values of $t$ and looking at predictions at configurations $\vtheta_0 + t \vv$. We repeat this for many random directions $\vv$. %
    \item %
    \textbf{Dropout subspace}: We start at an optimized solution $\vtheta_0$  apply dropout with a randomly chosen $p_\mathrm{keep}$, evaluate predictions at $\mathrm{dropout}_{p_\mathrm{keep}}(\vtheta_0)$ and repeat this many times with different  $p_\mathrm{keep}$ .
    \item \textbf{Diagonal Gaussian subspace}: We start at an optimized solution $\vtheta_0$ and look at the most recent iterations of training proceeding it. For each trainable parameter $\theta_i$, we calculate the mean $\mu_i$ and standard deviation $\sigma_i$ independently for each parameter, which corresponds to a diagonal covariance matrix.  This is similar to SWAG-diagonal \citep{maddox2019simple}. To sample solutions from the subspace, we repeatedly draw samples where each parameter independently as $\theta_i \sim \mathcal{N}(\mu_i,\sigma_i)$. %
   
    \item \textbf{Low-rank Gaussian subspace}: 
    We follow the same procedure as the diagonal Gaussian subspace above to compute the  mean $\mu_i$  for each trainable parameter. %
    For the covariance, we use a rank-$k$ approximation, by calculating the top $k$ principal components of the recent  weight vectors %
    $\{ {\vv}_i \in \mathbb{R}^\mathrm{params} \}_k$. We sample from a $k$-dimensional normal distribution and obtain the weight configurations as $\vtheta \sim {\vmu} + \sum_i \mathcal{N}(0^k,1^k) {\vv}_i$.  Throughout the text, we use the terms low-rank and PCA Gaussian interchangeably. %
\end{itemize}

\begin{figure}[ht]%
    \centering%
    \includegraphics[width=\sfigwidthtsne]{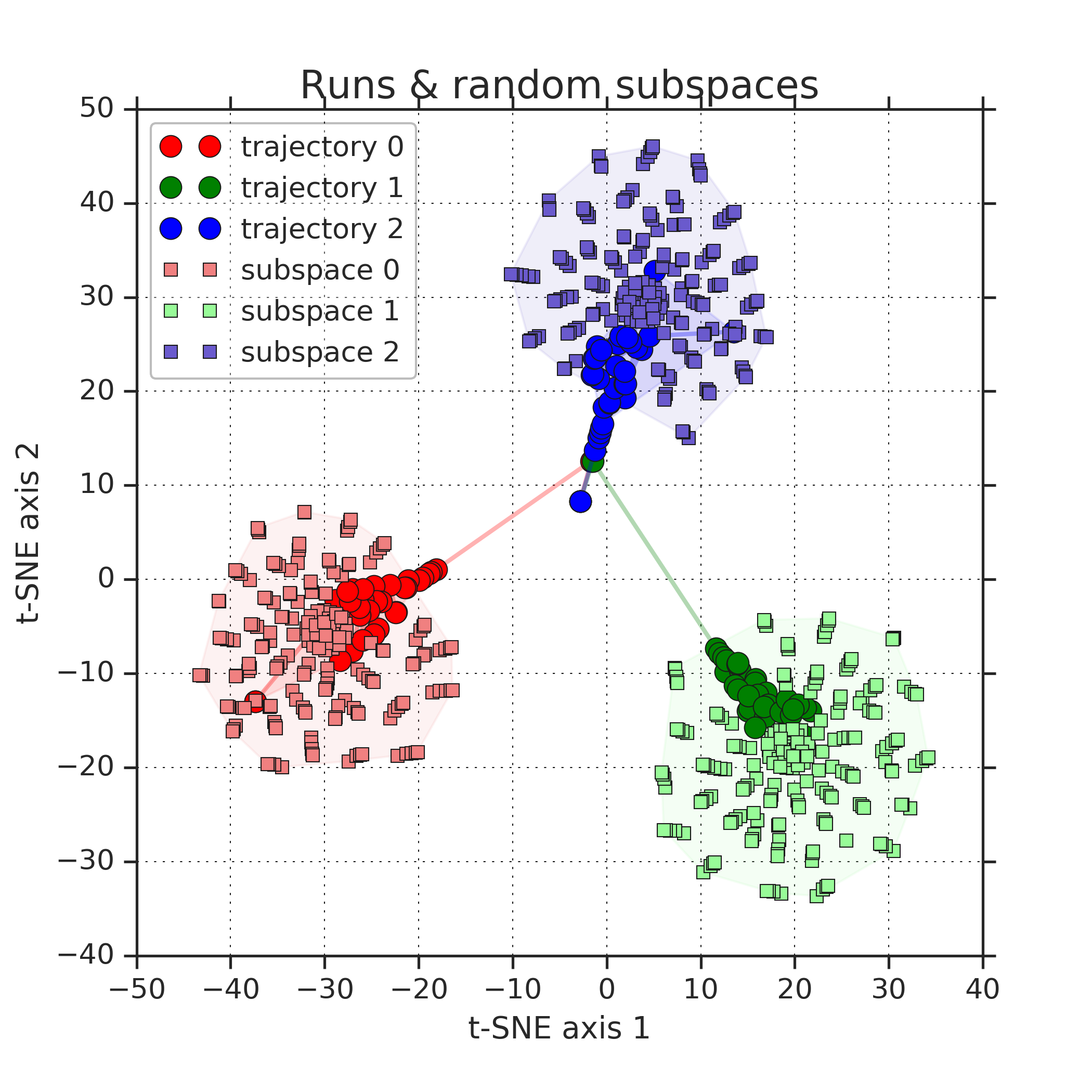}
    \includegraphics[width=\sfigwidthtsne]{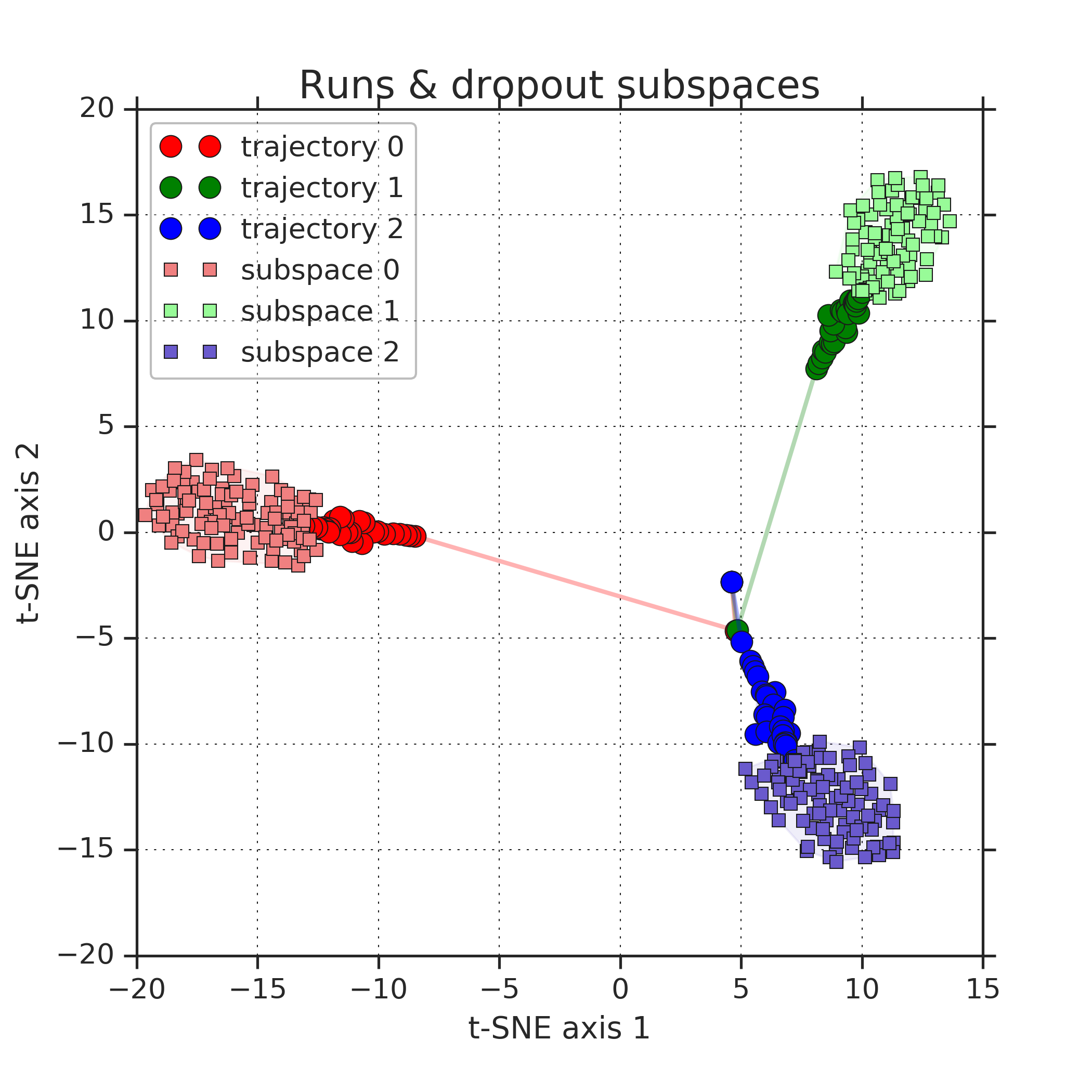} %
    \includegraphics[width=\sfigwidthtsne]{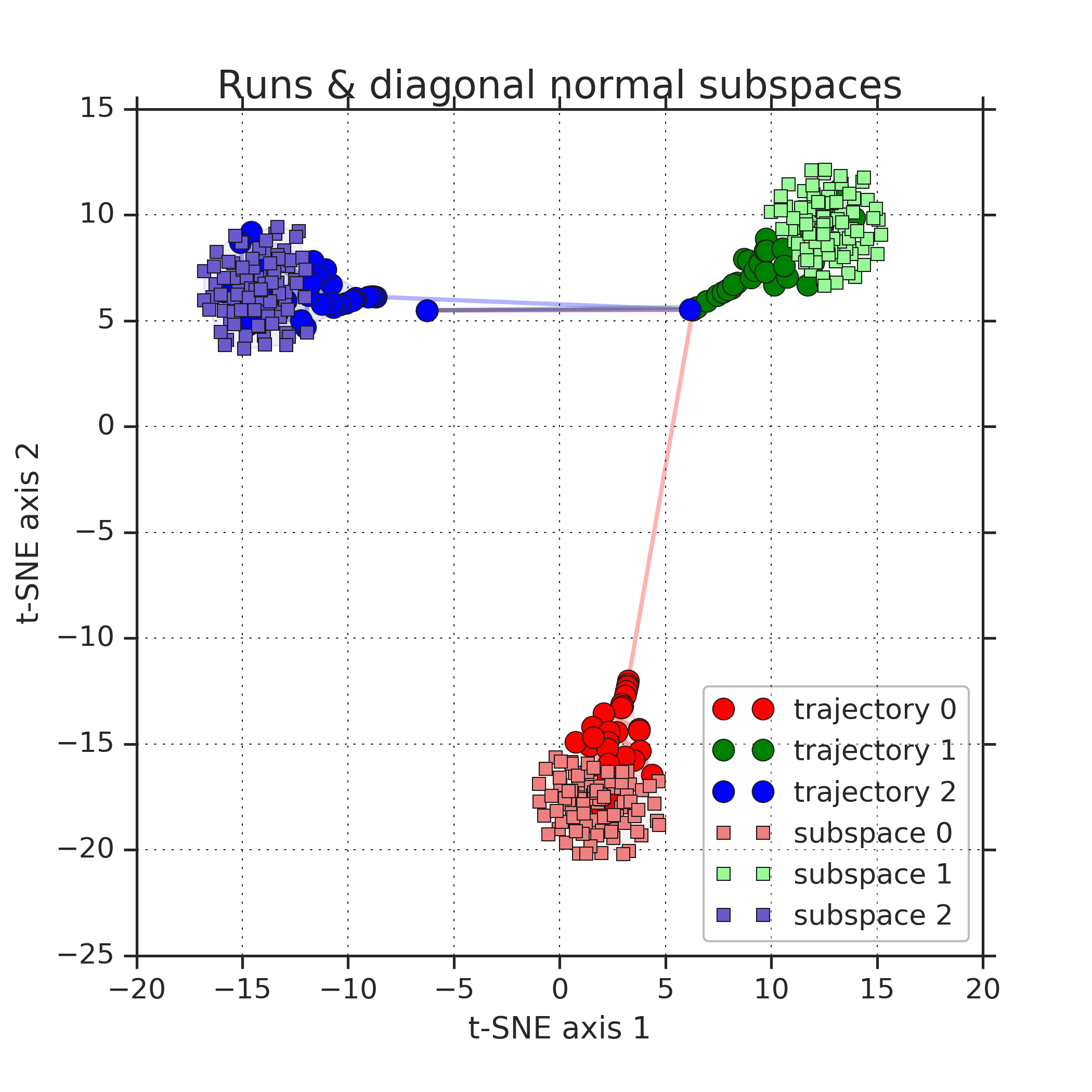}
    \includegraphics[width=\sfigwidthtsne]{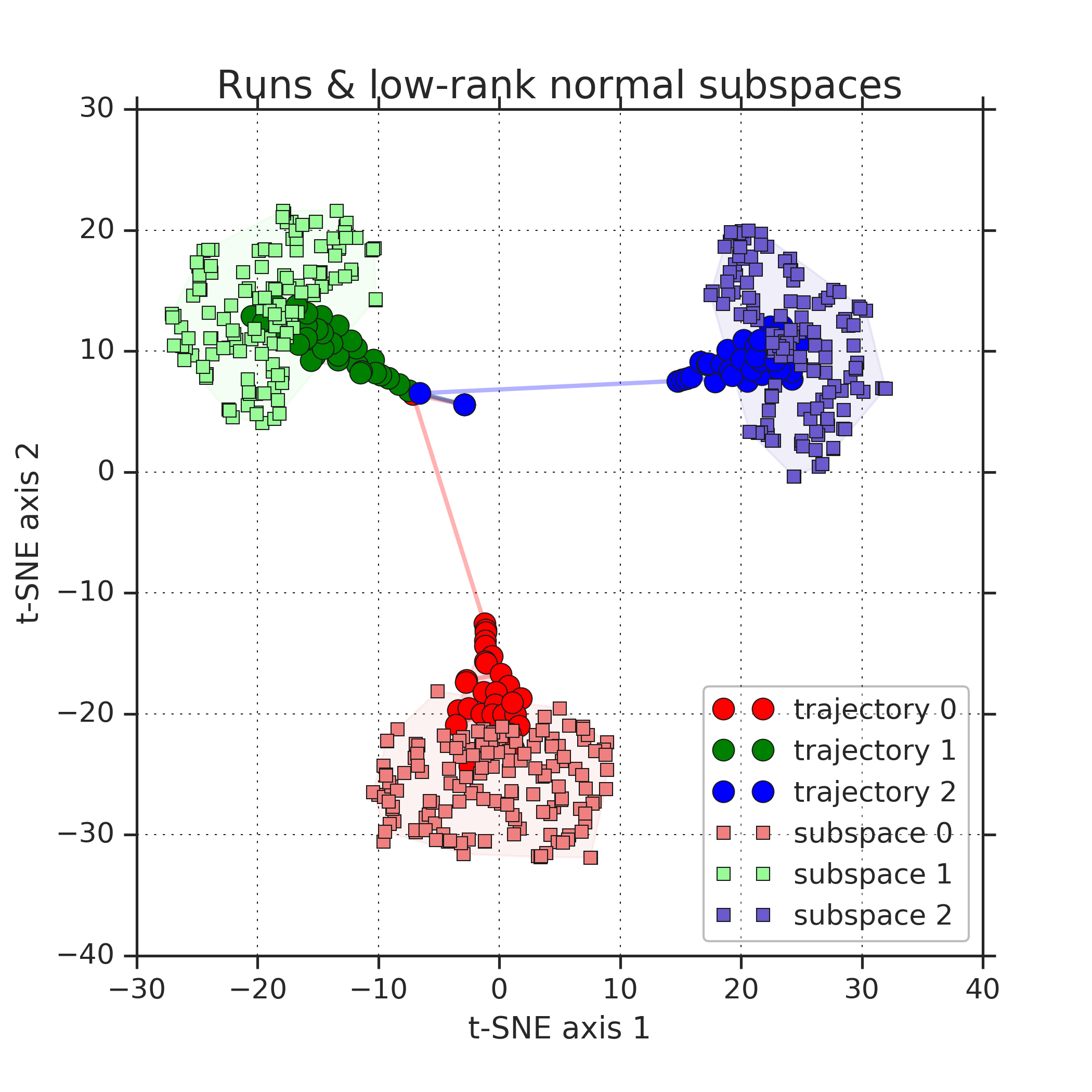}
     \reducespaceafterfigure
     \caption{Results using \textit{SmallCNN} on CIFAR-10: 
    t-SNE plots of validation set predictions for each trajectory along with four different subspace generation methods (showed by squares), in addition to 3 independently initialized and trained runs (different colors). As visible in the plot, the subspace-sampled functions stay in the prediction-space neighborhood of the run around which they were constructed, demonstrating that truly different functions are not sampled.
    }
    \label{fig:tSNE_subspace_plots}%
\end{figure}

Figure~\ref{fig:tSNE_subspace_plots} shows that functions sampled from a subspace (denoted by colored squares) corresponding to a particular initialization, are much more similar to each other. While some subspaces are more diverse, they still do not overlap with functions from another randomly initialized trajectory.  

\begin{figure}[ht]%
    \centering%
\includegraphics[width=\textwidth]{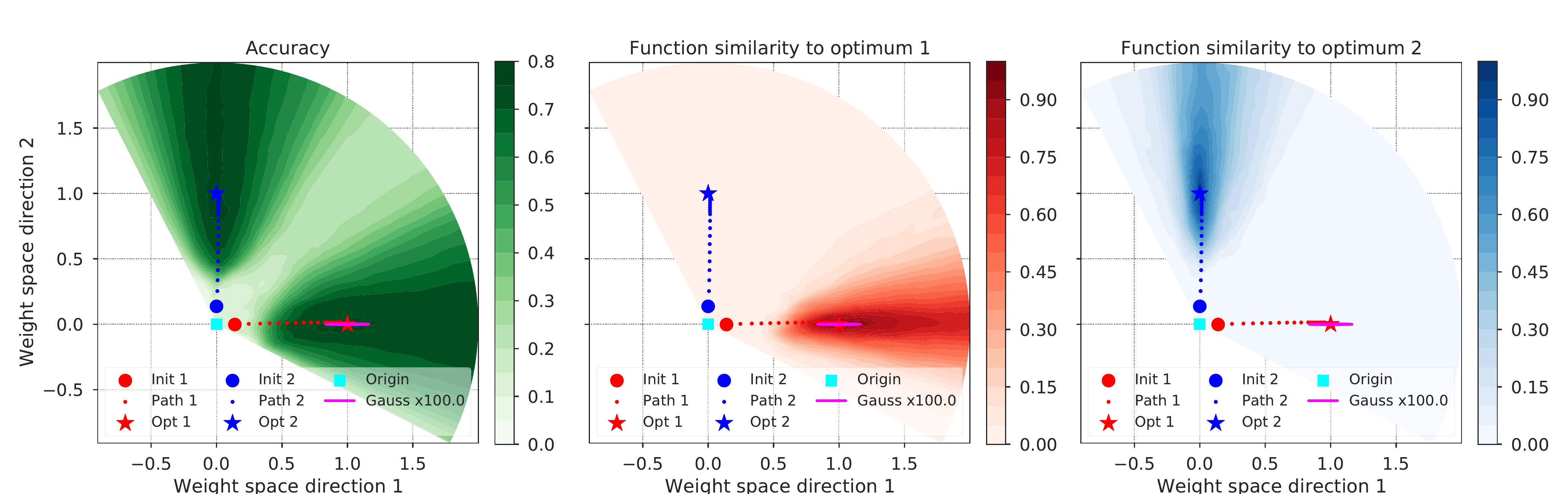}
     \reducespaceafterfigure
     \caption{\emph{Results using MediumCNN on CIFAR-10}: Radial loss landscape cut between the origin and two independent optima. %
     Left plot shows accuracy of models along the paths of the two independent trajectories, %
     and the middle and right plots show function space similarity to the two optima. }
    \label{fig:CNN_radial_cuts_trajectory}%
\end{figure}%

As additional evidence, Figure~\ref{fig:CNN_radial_cuts_trajectory} provides a two-dimensional visualization of the radial landscape along the directions of two different optima. The 2D sections of the weight space visualized are defined by the origin (all weights are 0) and two independently initialized and trained optima. The weights of the two trajectories (shown in red and blue) are initialized using standard techniques and they increase radially with training due to their softmax cross-entropy loss. The left subplot shows that different randomly initialized trajectories eventually achieve similar accuracy. We also sample from a Gaussian subspace along trajectory 1 (shown in pink). 
The middle and the right subplots show function space similarity (defined as the fraction of points on which they agree on the class prediction) of the parameters along the path to optima 1 and 2.  Solutions along each trajectory (and Gaussian subspace) are much more similar to their respective optima, which is consistent with the cosine similarity and t-SNE plots.  %

 \subsection{Diversity versus Accuracy plots} 

 To illustrate  the difference in another fashion, 
 we sample functions from a single subspace and plot accuracy versus diversity, as measured by disagreement between predictions from the baseline solution. From a bias-variance trade-off perspective, we require a procedure to produce functions that are accurate (which leads to low bias by aggregation) as well as de-correlated (which leads to lower variance by aggregation). Hence, the diversity vs accuracy plot allows us to visualize the trade-off that can be achieved by subspace sampling methods versus deep ensembles.  %

The diversity score quantifies the difference of two functions (a base solution and a sampled one), by measuring fraction of  datapoints on which their predictions differ. 
We chose this approach due to its simplicity; we also computed the KL-divergence and other distances between the output probability distributions, leading to equivalent conclusions.  %
Let $d_\mathrm{diff}$ denote the fraction of predictions on which the two functions differ. It is $0$ when the two functions make identical class predictions, and $1$ when they differ on every single example. To account for the fact that the lower the accuracy of a function, the higher its potential $d_\mathrm{diff}$ due to the possibility of the wrong answers being random and uncorrelated between the two functions, we normalize this by $(1 - a)$, where $a$ is the accuracy of the sampled solution. We also derive idealized lower and upper limits of these curves (showed in dashed lines) by perturbing the reference solution's predictions (lower limit) and completely random predictions at a given accuracy (upper limit), see Appendix~\ref{sec:limit:curves} for a discussion. 

\begin{figure*}[ht]%
    \centering%
    \includegraphics[width=0.32\linewidth]{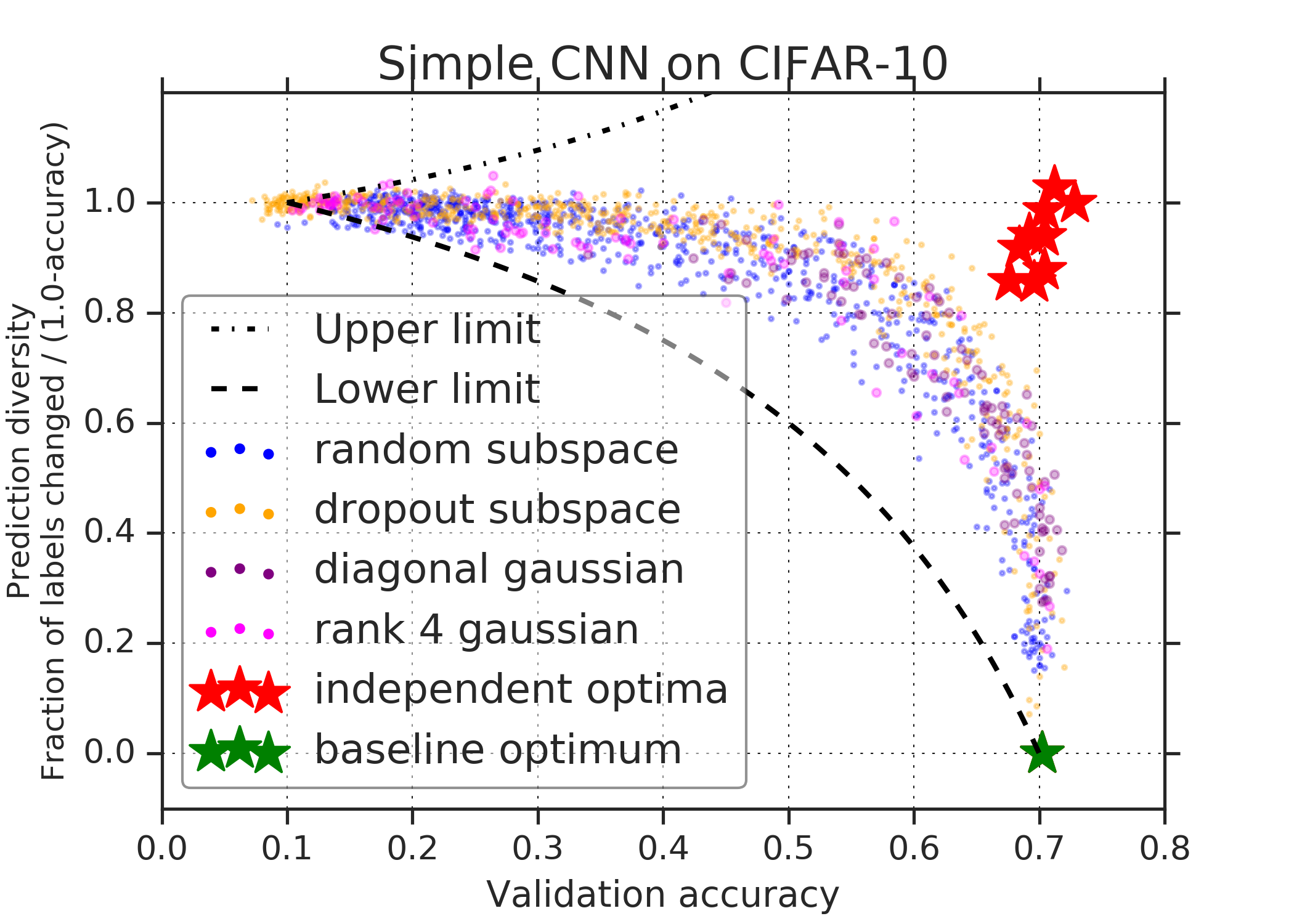}
    \includegraphics[width=0.32\linewidth]{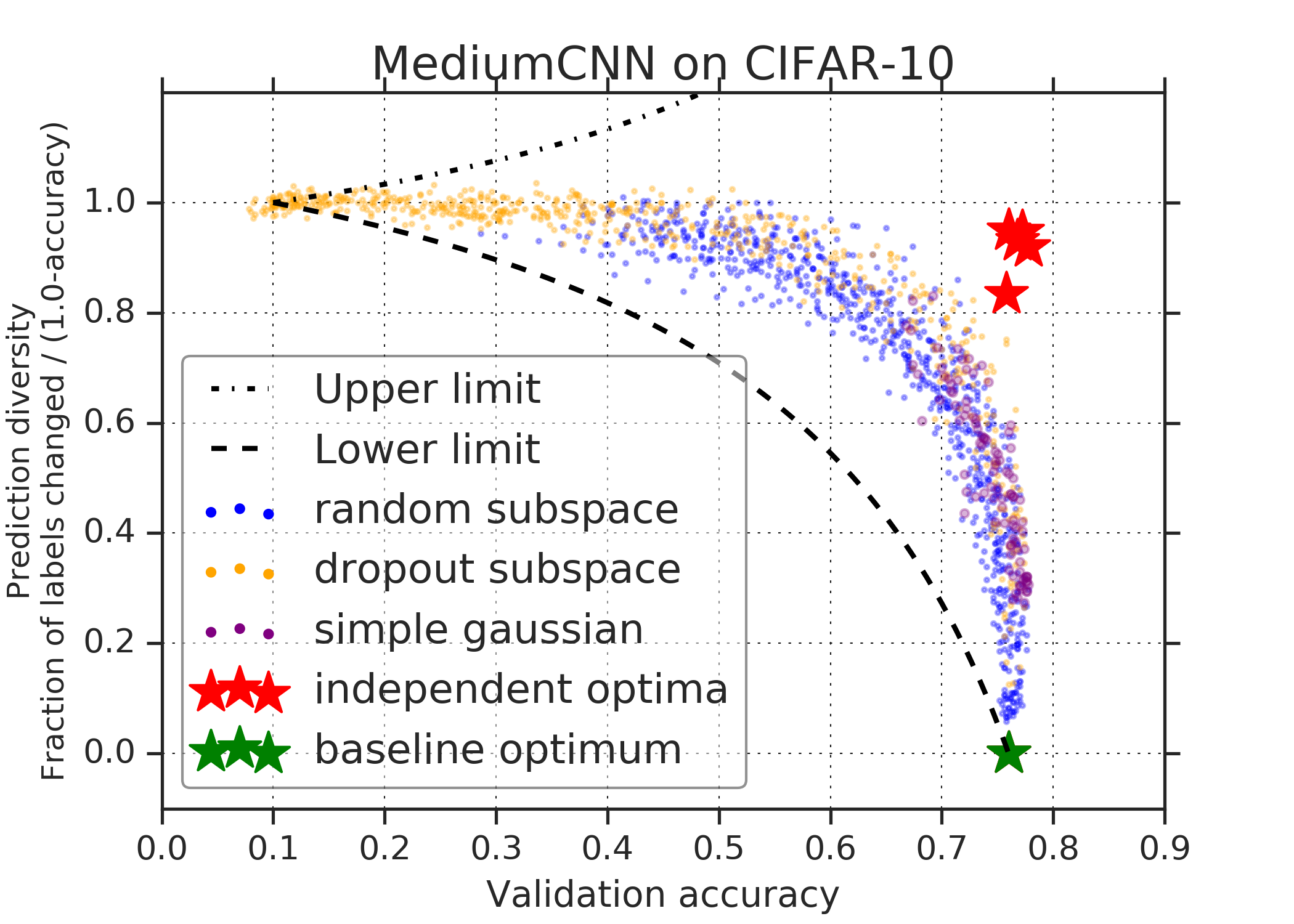}
    \includegraphics[width=0.32\linewidth]{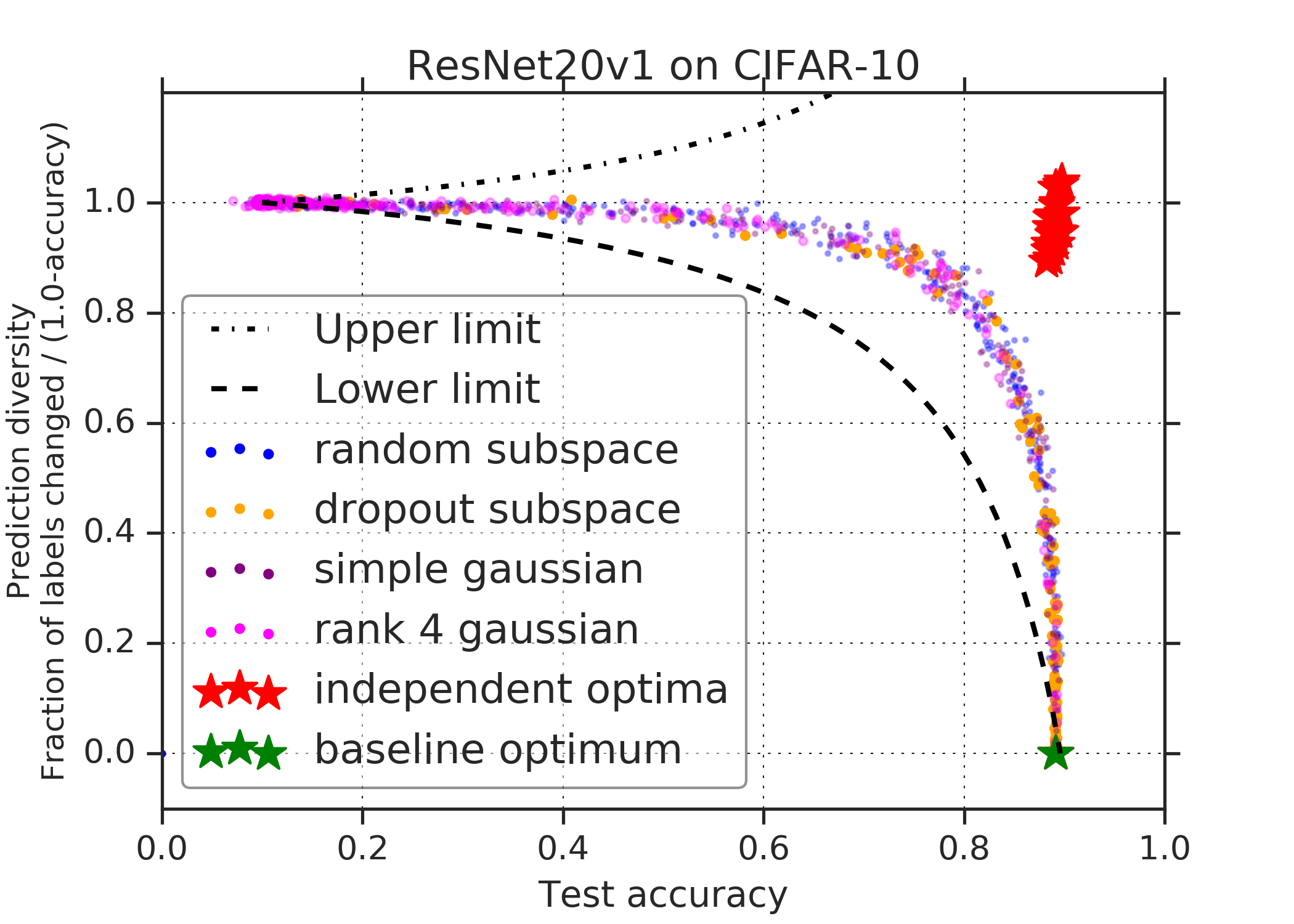}
    \reducespaceafterfigure %
      \caption{
    Diversity versus accuracy plots for 3 models trained on CIFAR-10: \emph{SmallCNN}, \emph{MediumCNN} and a \emph{ResNet20v1}. Independently initialized and optimized solutions (red stars) achieve better diversity vs accuracy trade-off than the four different subspace sampling methods.
    }
    \label{fig:diversity_plots_all}%
\end{figure*}

\begin{figure}[ht]
\begin{center}
   \begin{subfigure}[\emph{ResNet20v1} trained on CIFAR-100.]{ 
      \includegraphics[width=0.35\linewidth]{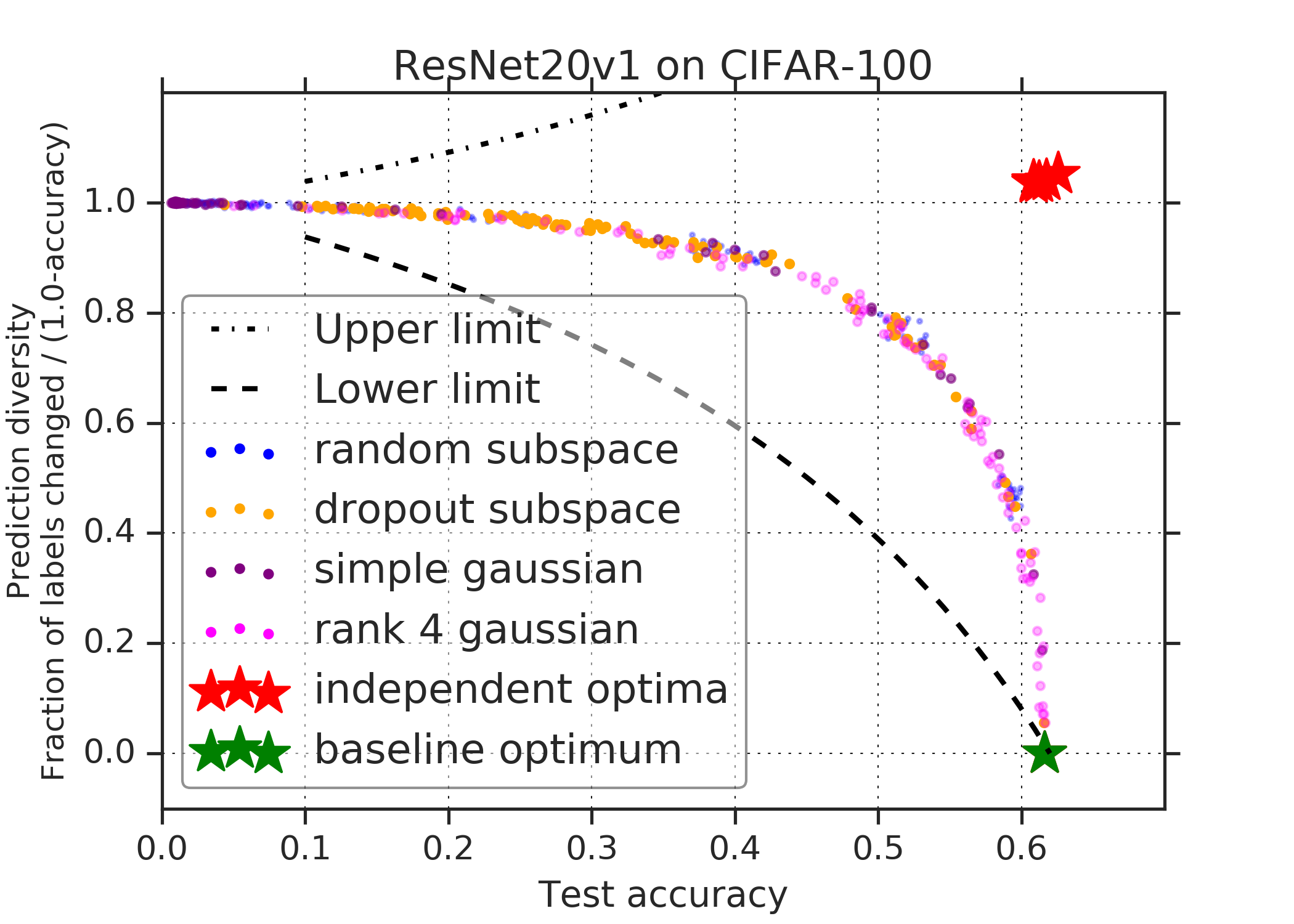}
       \label{fig:diversity_accuracy_ResNet_CIFAR100}%
   }\end{subfigure}
    \begin{subfigure}[\emph{ResNet50v2} trained on ImageNet.]{ \includegraphics[width=0.35\textwidth]{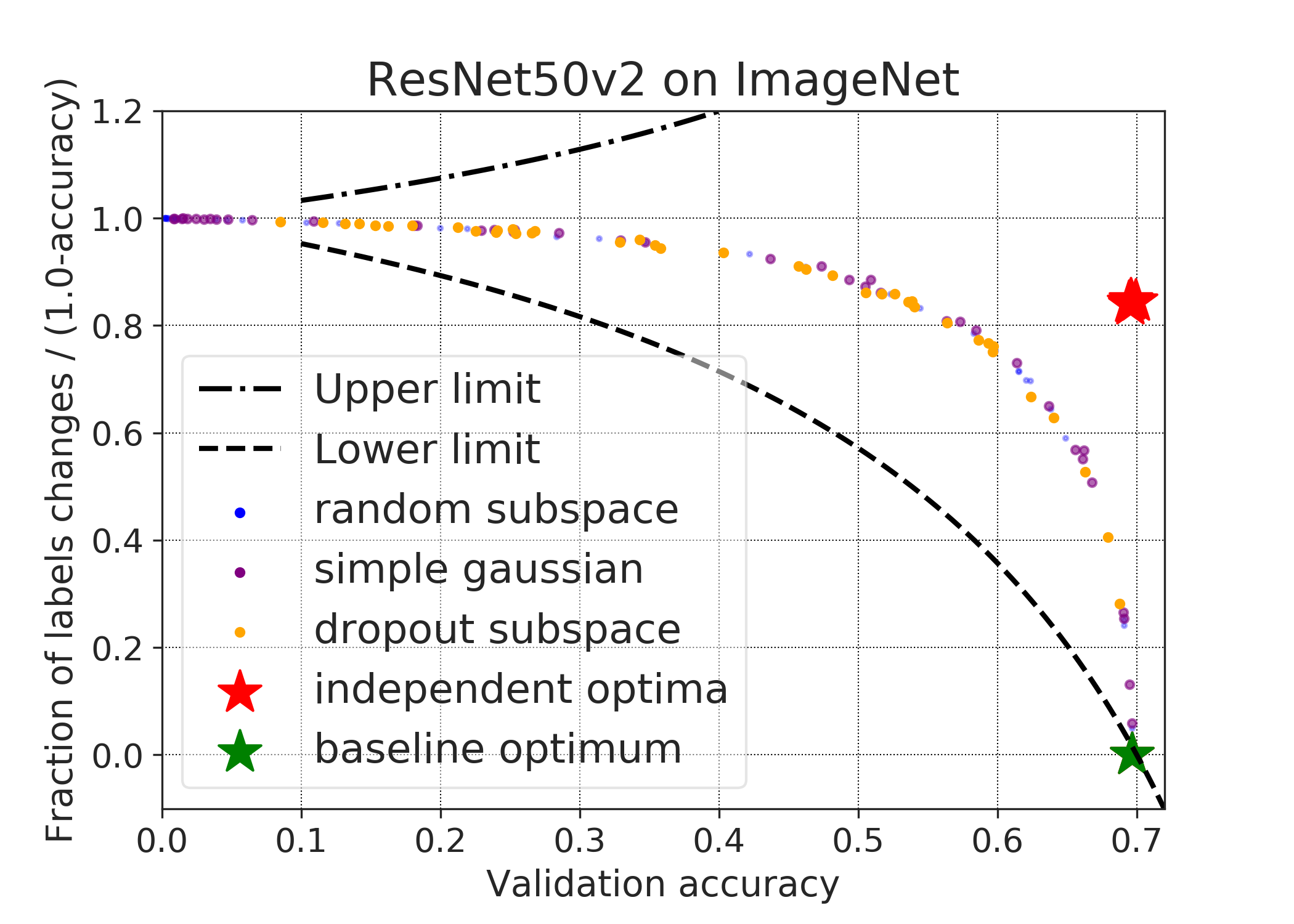}
     \label{fig:diversity_accuracy_ResNet_ImageNet}%
    }\end{subfigure}
    \reducespaceafterfigure
      \caption{Diversity vs. 
    accuracy plots for \emph{ResNet20v1}  on CIFAR-100, and \emph{ResNet50v2}  on ImageNet. %
      }
    \label{fig:diversity_accuracy_ResNet_CIFAR100_ImageNet}%
      \end{center}
\end{figure}

Figure~\ref{fig:diversity_plots_all} shows the results on CIFAR-10. %
Comparing these subspace points (colored dots) to the baseline optima (green star) and the optima from different random initializations (denoted by red stars), we observe that random initializations are much more effective at sampling diverse and accurate solutions, than subspace based methods constructed out of a single trajectory. The results are consistent across different architectures and datasets. Figure~\ref{fig:diversity_accuracy_ResNet_CIFAR100_ImageNet} shows results on CIFAR-100 and ImageNet. We observe that solutions obtained by subspace sampling methods have a worse trade off between accuracy and prediction diversity, 
compared to independently initialized and trained optima. Interestingly, the separation between the subspace sampling methods and independent optima in the diversity--accuracy plane gets more pronounced the more difficult the problem and the more powerful the network.    

\vspace{-1em}
\section{%
Evaluating the Relative Effects of Ensembling versus Subspace Methods
}
\label{sec:subspaces}

Our hypothesis in Figure~\ref{fig:ensemble-vs-bayes} and the empirical observations in the previous section suggest that subspace-based methods and ensembling should provide complementary benefits in terms of uncertainty and accuracy. %
Since our goal is not to propose a new method, but to carefully test this hypothesis, we
evaluate the performance of the following four variants for controlled comparison: %
\begin{itemize}\itemsep0em
\item \emph{Baseline}: optimum at the end of a single  trajectory.
\item \emph{Subspace sampling}: average predictions over the solutions sampled from a subspace.
\item \emph{Ensemble}: train baseline multiple times with random initialization and average the predictions.
\item \emph{Ensemble + Subspace sampling}: train multiple times with random initialization, and use subspace sampling within each trajectory.
\end{itemize}

{To maintain the accuracy of random samples at a reasonable level for fair comparison, we reject the sample if validation accuracy is below $0.65$. For the CIFAR-10 experiment, we  use a rank-4 approximation of the random samples using PCA.} 
Note that diagonal Gaussian, low-rank Gaussian and random subspace sampling methods to approximate each mode of the posterior leads to an increase in the number of parameters required for each mode.  However, using just the mean weights for each mode would not cause such an increase.  \citet{izmailov2018averaging} proposed stochastic weight averaging (SWA) for better generalization.   
One could also compute an (exponential moving) average of the weights along the trajectory, inspired by Polyak-Ruppert averaging in convex optimization, (see also \citep{mandt2017stochastic} for a Bayesian view on iterate averaging). As weight averaging (WA) has been already studied by \citet{izmailov2018averaging}, we do not discuss it in detail. 
Our goal is to test if WA finds a better point estimate within each mode (see cartoon illustration in Figure~\ref{fig:ensemble-vs-bayes}) and
provides complementary benefits to ensembling over random initialization. %
In our experiments, we use WA on the last few epochs which corresponds to using just the mean of the parameters within each mode.   

\begin{figure}[ht]%
    \centering%
    \reducespaceafterfigure
      \includegraphics[width=\sfigwidthtwo]{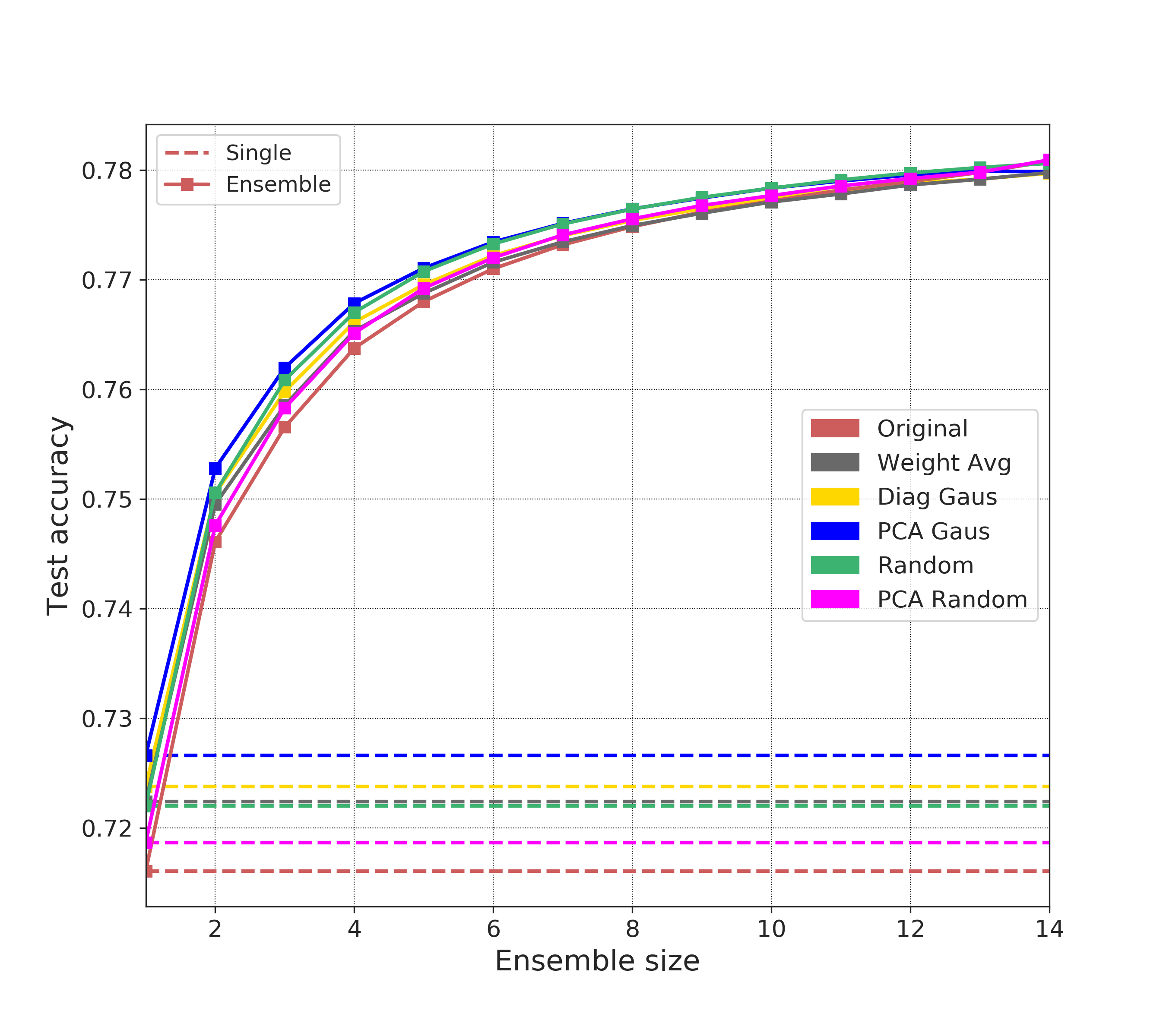}%
       \includegraphics[width=\sfigwidthtwo]{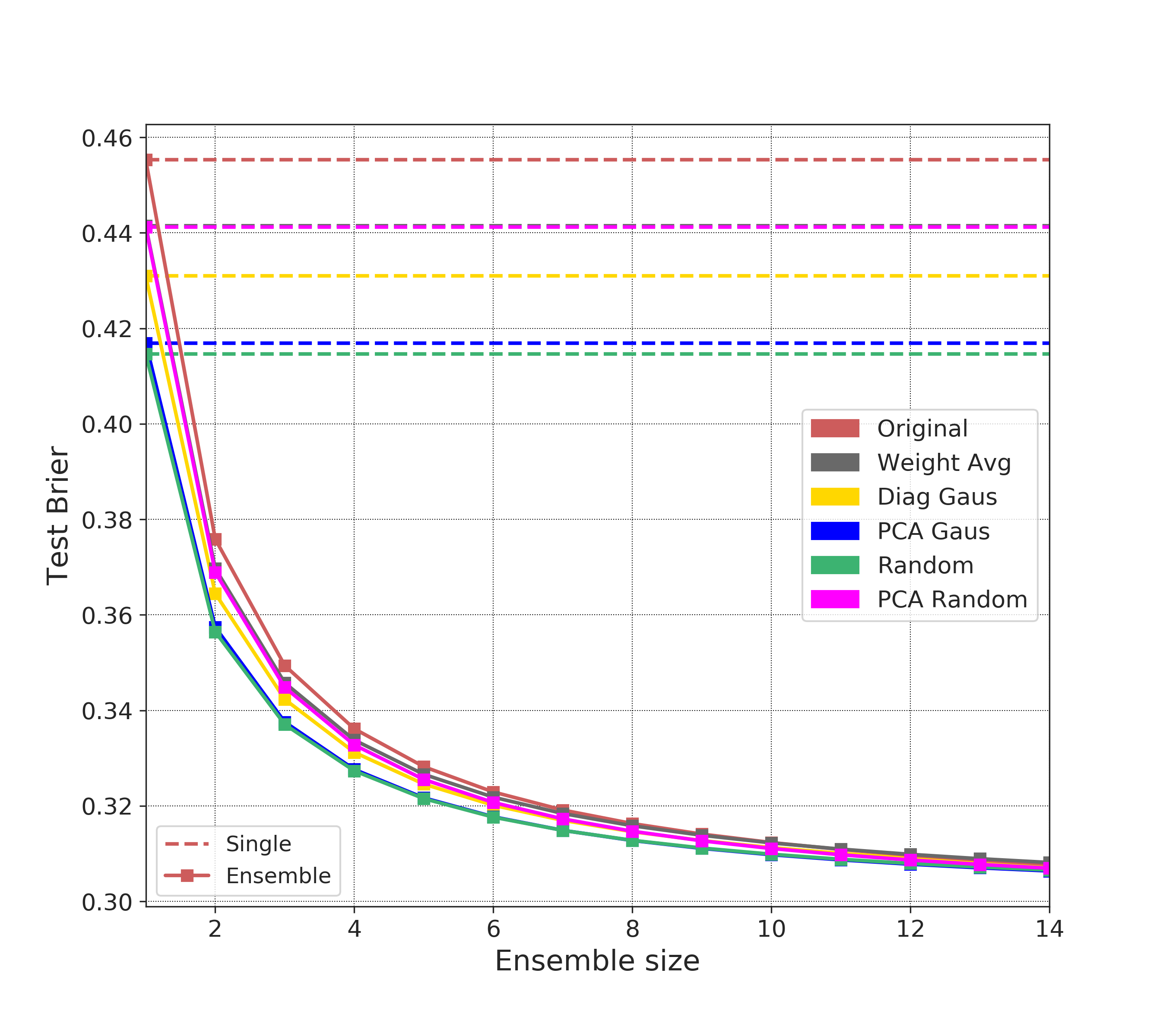}%
  \reducespaceafterfigure   \caption{
    Results using MediumCNN on CIFAR-10 showing the  complementary benefits of ensemble and subspace methods as a function of ensemble size. We used 10 samples for each subspace method.
    }
    \label{fig:ensemble-subspace-cifar}%
\end{figure}

Figure~\ref{fig:ensemble-subspace-cifar} shows the results on CIFAR-10.  
The results validate our hypothesis that (i) subspace sampling and ensembling  provide complementary benefits, and (ii) the relative benefits of ensembling are higher as it averages predictions over more diverse solutions.  

\begin{figure}[th]%
    \centering%
 \includegraphics[width=\sfigwidththreesubfigure]{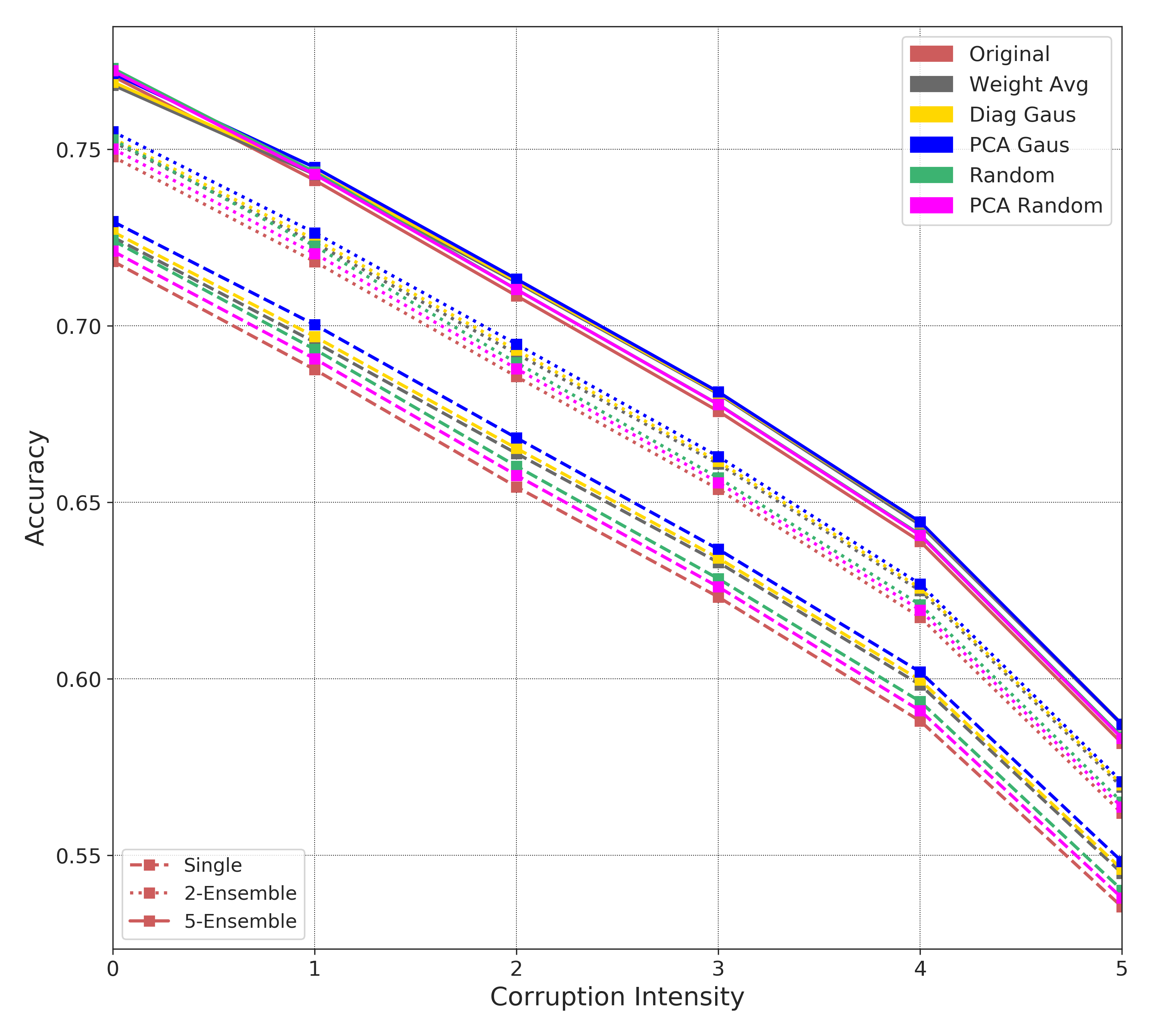}%
       \includegraphics[width=\sfigwidththreesubfigure]{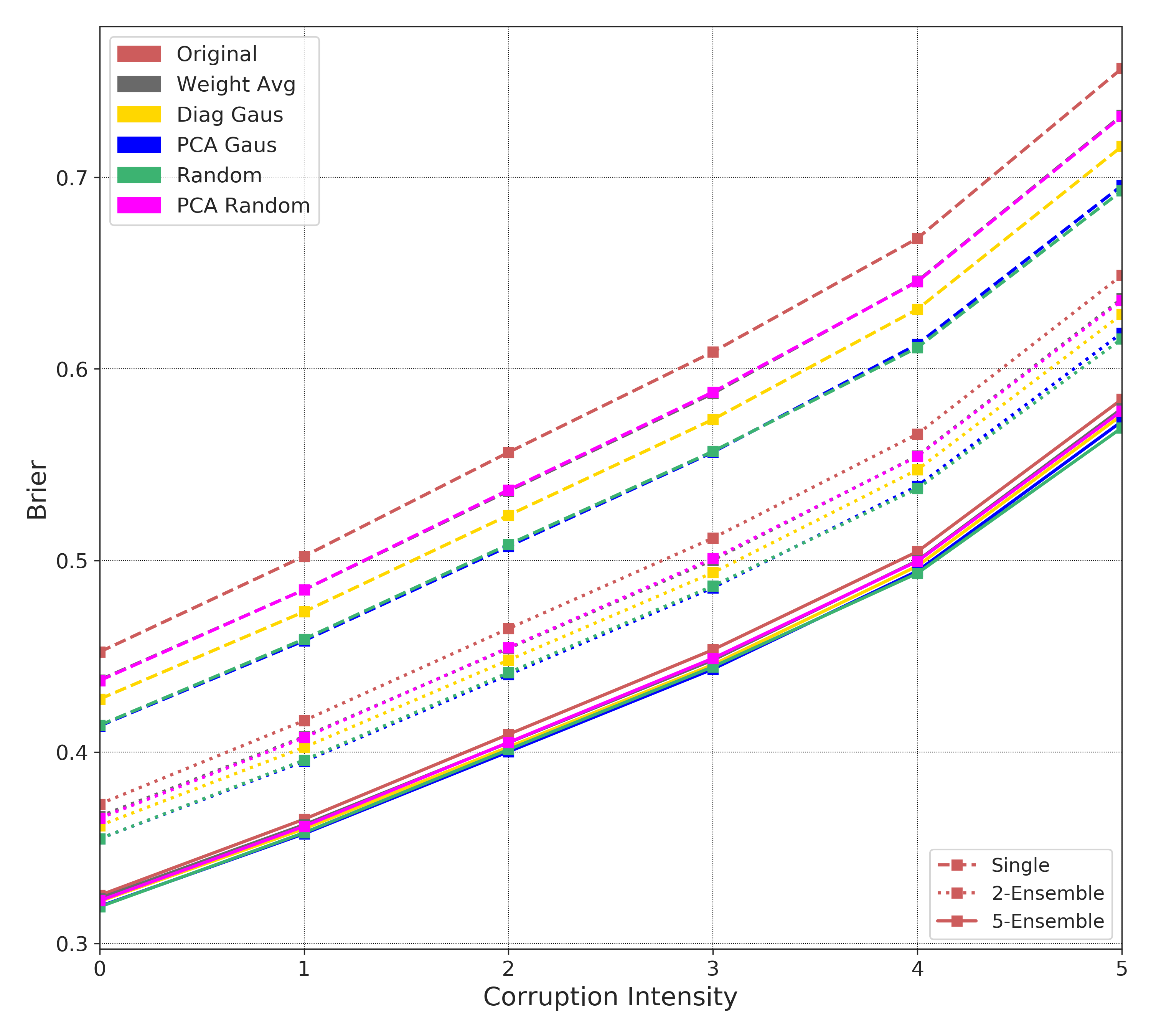}%
       \includegraphics[width=\sfigwidththreesubfigure]{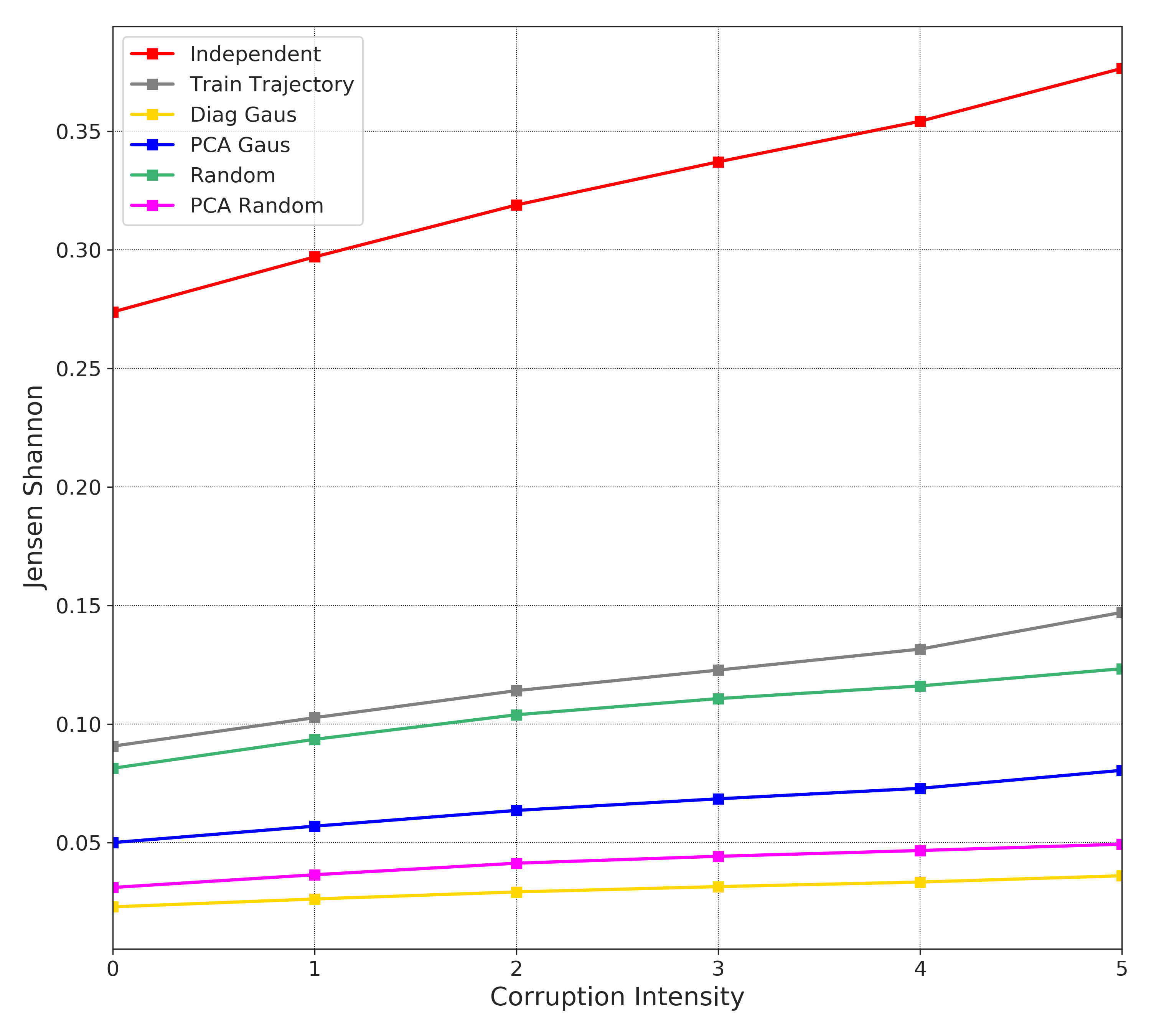}
   \reducespaceafterfigure %
    \reducespaceafterfigure
    \caption{
   Results using \emph{MediumCNN} on CIFAR-10-C for varying levels of corruption intensity. Left plot shows accuracy, medium plot shows Brier score and right plot shows Jensen-Shannon divergence. %
    }
      \label{fig:cifar10:acc-brier:clean-corruptions}%
\end{figure}

\textbf{Effect of function space diversity on dataset shift} 
We test the same hypothesis under dataset shift \citep{ovadia2019can,hendrycks2018benchmarking}.
Left and middle subplots of  Figure~\ref{fig:cifar10:acc-brier:clean-corruptions}
show accuracy and Brier score  on the CIFAR-10-C benchmark. We observe again that ensembles and subspace sampling methods provide complementary benefits. 

The diversity versus accuracy plot compares diversity to a reference solution, but it is also important to also look at the diversity between \emph{multiple samples of the same method}, as this will effectively determine the efficiency of the method in terms of the bias-variance trade-off. Function space diversity is particularly important  to avoid overconfident predictions under dataset shift, as averaging over similar functions would not reduce overconfidence. To visualize this, we draw 5 samples of each method and compute the average Jensen-Shannon divergence between their predictions, defined as 
$\sum_{m=1}^M KL(p_{\vtheta_m}(y|\vx)||\bar{p}(y|\vx))$ where $KL$ denotes the Kullback-Leibler divergence and $\bar{p}(y|\vx)=(1/M) \sum_{m} p_{\vtheta_m}(y|\vx)$.
Right subplot of Figure~\ref{fig:cifar10:acc-brier:clean-corruptions}
shows the results on CIFAR-10-C for increasing corruption intensity. We observe that Jensen-Shannon divergence is the highest between independent random initializations, and lower for subspace sampling methods; the difference is higher under dataset shift, which explains the findings of \citet{ovadia2019can} that deep ensembles outperform other methods under dataset shift. %
{We also observe similar trends when testing on an OOD dataset such as SVHN: JS divergence is 0.384 for independent runs, 0.153 for within-trajectory, 0.155 for random sampling, 
0.087 for rank-5 PCA Gaussian and 0.034  for diagonal Gaussian.
 } 
 
\begin{figure}[ht]%
    \centering%

\includegraphics[width=1.0\linewidth]{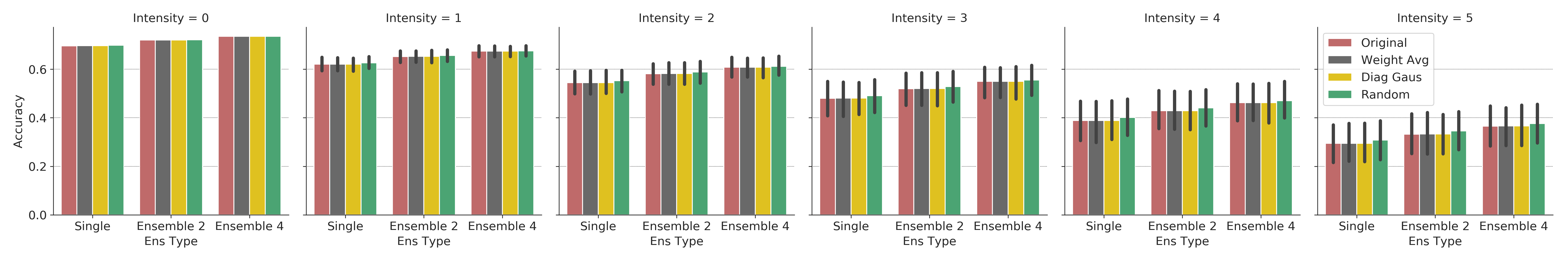}\\
    
\includegraphics[width=1.0\linewidth]{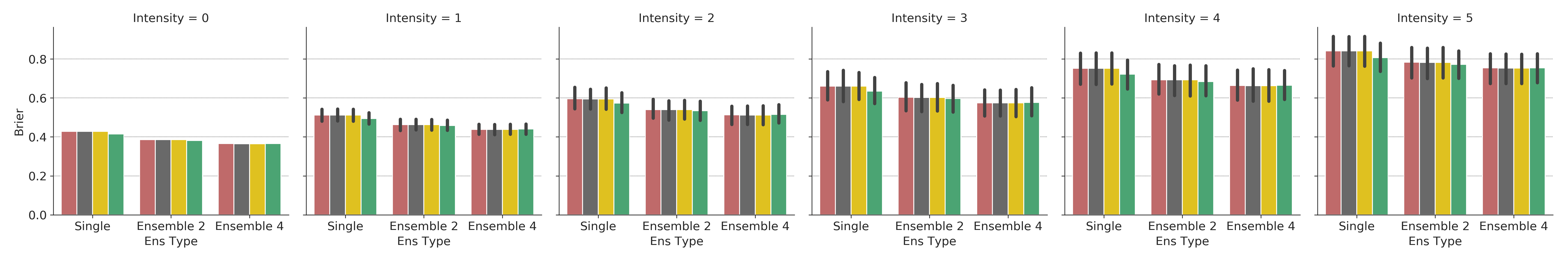}
  \reducespaceafterfigure %
  \reducespaceafterfigure
    \caption{Results using \emph{ResNet50v2} on ImageNet   test and ImageNet-C for varying corruptions. %
    }
    \label{fig:imagenet:acc-brier:clean-corruptions}%
\end{figure}

 \textbf{Results on ImageNet} 
To illustrate the effect on another challenging dataset, we repeat these experiments on ImageNet \citep{imagenet_cvpr09} using \emph{ResNet50v2} architecture. Due to computational constraints, we do not evaluate PCA subspace on ImageNet.  %
Figure~\ref{fig:imagenet:acc-brier:clean-corruptions} shows results on ImageNet test set (zero corruption intensity) and ImageNet-C for increasing corruption intensities.  Similar to CIFAR-10, random subspace performs best within subspace sampling methods, and provides complementary benefits to random initialization.  
We empirically observed that the relative gains of WA (or subspace sampling) are smaller when the individual models 
converge to a better optima within each mode. Carefully choosing which points to average, e.g. using cyclic learning rate as done in fast geometric ensembling \citep{garipov2018loss} can yield further benefits.

\section{Discussion} 
Through extensive experiments, we show that trajectories of randomly initialized neural networks explore different modes in function space, which explains why deep ensembles trained with just random initializations work well in practice. %
Subspace sampling methods such as weight averaging, Monte Carlo dropout, and various versions of local Gaussian approximations, sample functions that might lie relatively far from the starting point in the weight space, however, they remain similar %
in function space, giving rise to an insufficiently diverse set of predictions. Using the concept of the diversity--accuracy plane, we demonstrate empirically that current variational Bayesian methods do not reach the trade-off between diversity and accuracy achieved by independently trained models. %
There are several interesting directions for future research: understanding the role of random initialization on training dynamics (see Appendix~\ref{sec:randomness} for a preliminary investigation), exploring methods which achieve higher diversity than deep ensembles (e.g.~through explicit decorrelation), and developing parameter-efficient  methods (e.g. implicit ensembles or Bayesian deep learning algorithms) that achieve better diversity--accuracy trade-off than deep ensembles. %

\clearpage
\newpage

\bibliography{main}
\bibliographystyle{unsrtnat}

 \clearpage\newpage
 \appendix
 {\begin{center} {\Large{\textbf{Supplementary Material}}}
\end{center}}
 \setcounter{figure}{0}
\setcounter{table}{0}
\makeatletter 
\renewcommand{\thefigure}{S\@arabic\c@figure}
\renewcommand{\thetable}{S\@arabic\c@table}
\makeatother

\section{Identical loss does not imply identical functions in prediction space}
\label{sec:diversity-along-tunnel}
\begin{figure}[ht]%
    \centering%
           \begin{subfigure}[Accuracy along the radial loss-landscape cut]{ 
             \includegraphics[width=\sfigwidthlandscape]{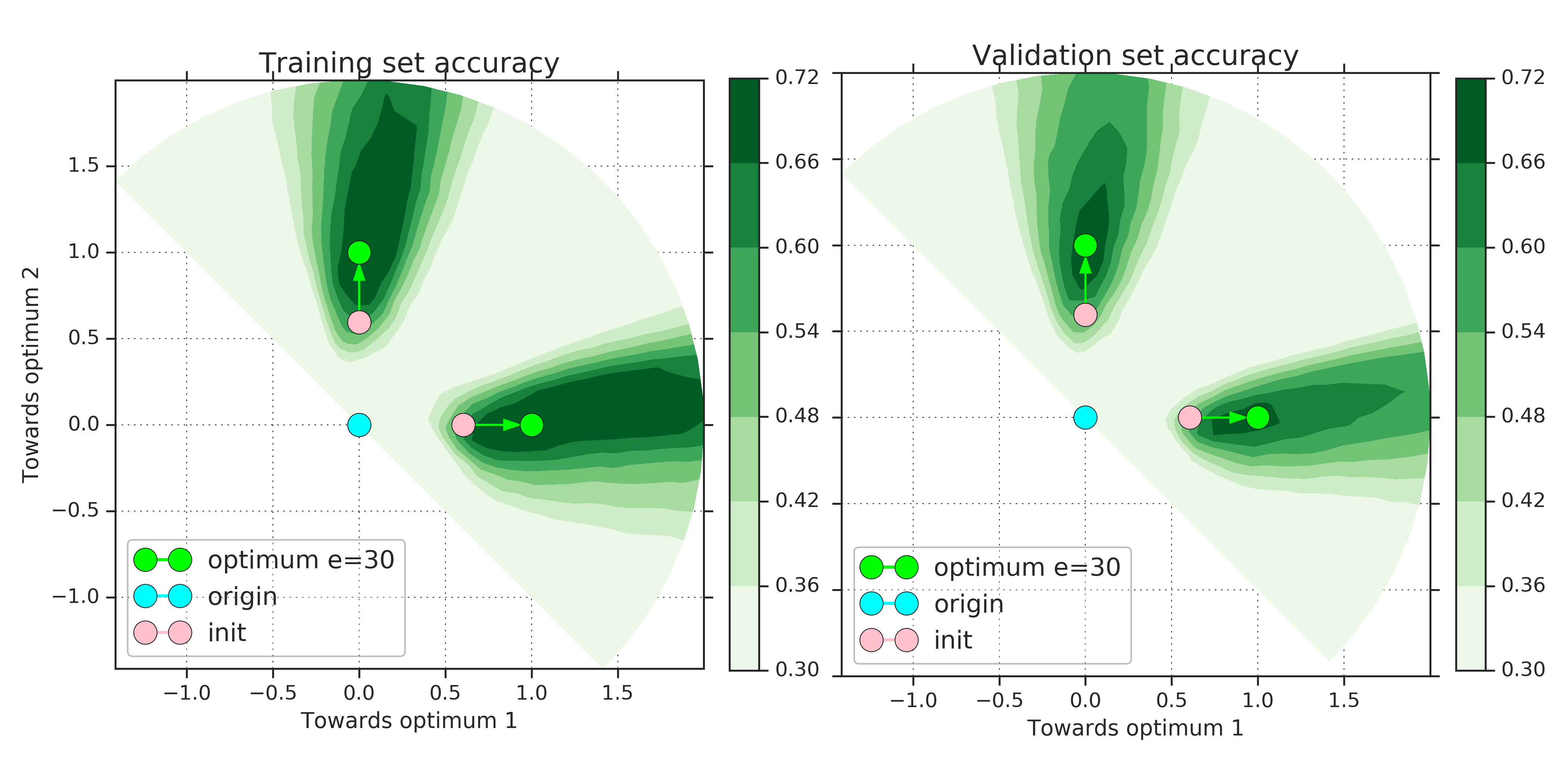}
                    }\end{subfigure}
                \begin{subfigure}[Function-space similarity]{ 
         \includegraphics[width=\sfigwidthlandscape]{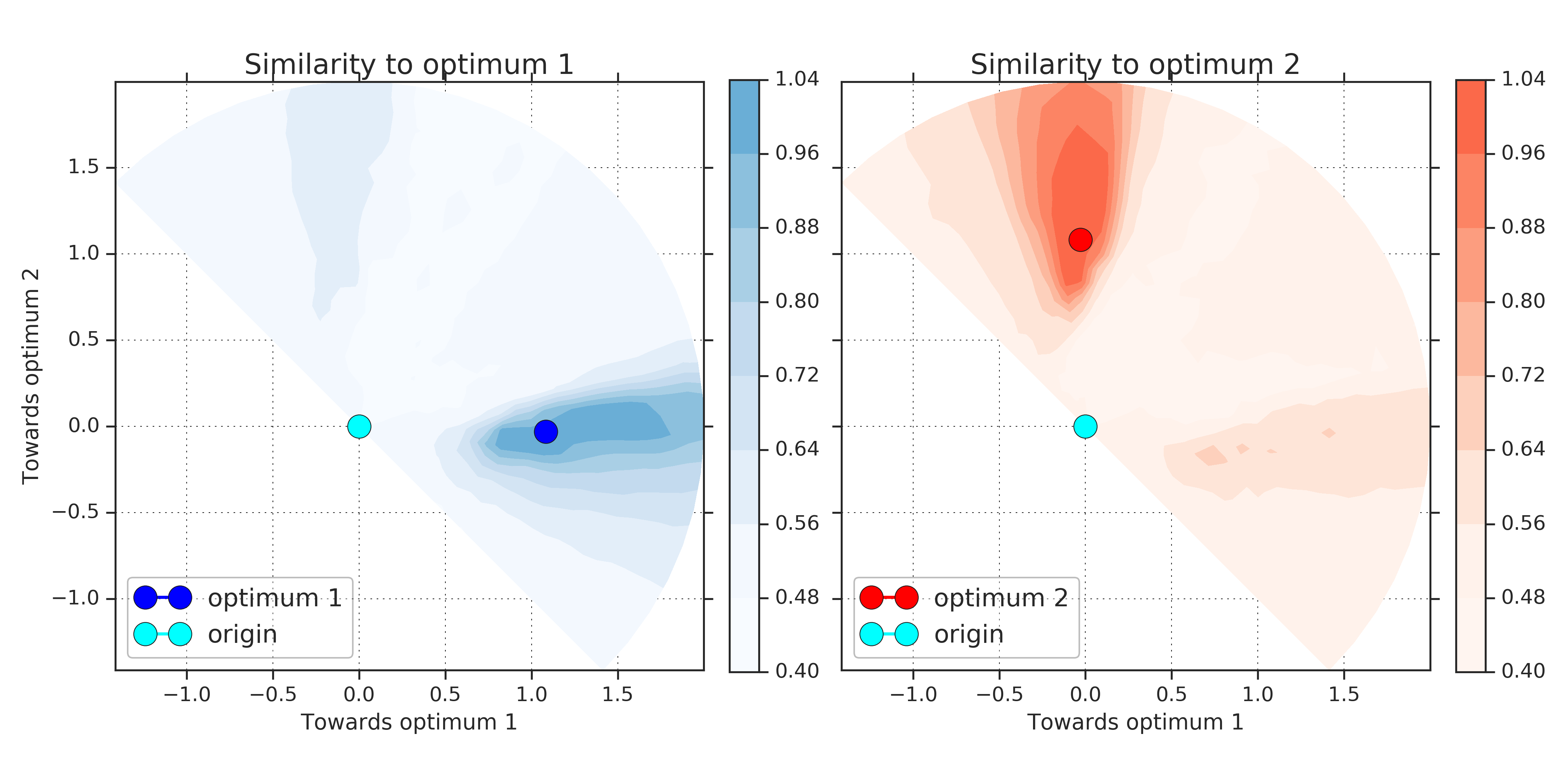}
                }\end{subfigure}  
     \reducespaceafterfigure
     \caption{\emph{Results using MediumCNN on CIFAR-10}: Radial loss landscape cut between the origin and two independent optima and the predictions of models on the same plane.}
    \label{fig:SimpleCNN_radial_cuts}%
\end{figure}

To make our paper self-contained, we review the literature on loss surfaces and mode connectivity \citep{garipov2018loss,draxler2018essentially,fort2019large}. We provide visualizations of the loss surface which confirm the findings of prior work as well as complement them.  
Figure~\ref{fig:SimpleCNN_radial_cuts} shows the radial loss landscape (train as well as the validation set) along the directions of two different optima. %
The left subplot shows that different trajectories achieve similar values of the loss, and the right subplot shows the similarity of these functions to their respective optima  (in particular the fraction of labels predicted on which they differ divided by their error rate). While the loss values from different optima are similar, the functions are different, which confirms that random initialization leads to different modes in function space. %

\begin{figure}[ht]
\begin{center}
   \begin{subfigure}[Cartoon illustration]{  
  \includegraphics[width=0.31\linewidth]{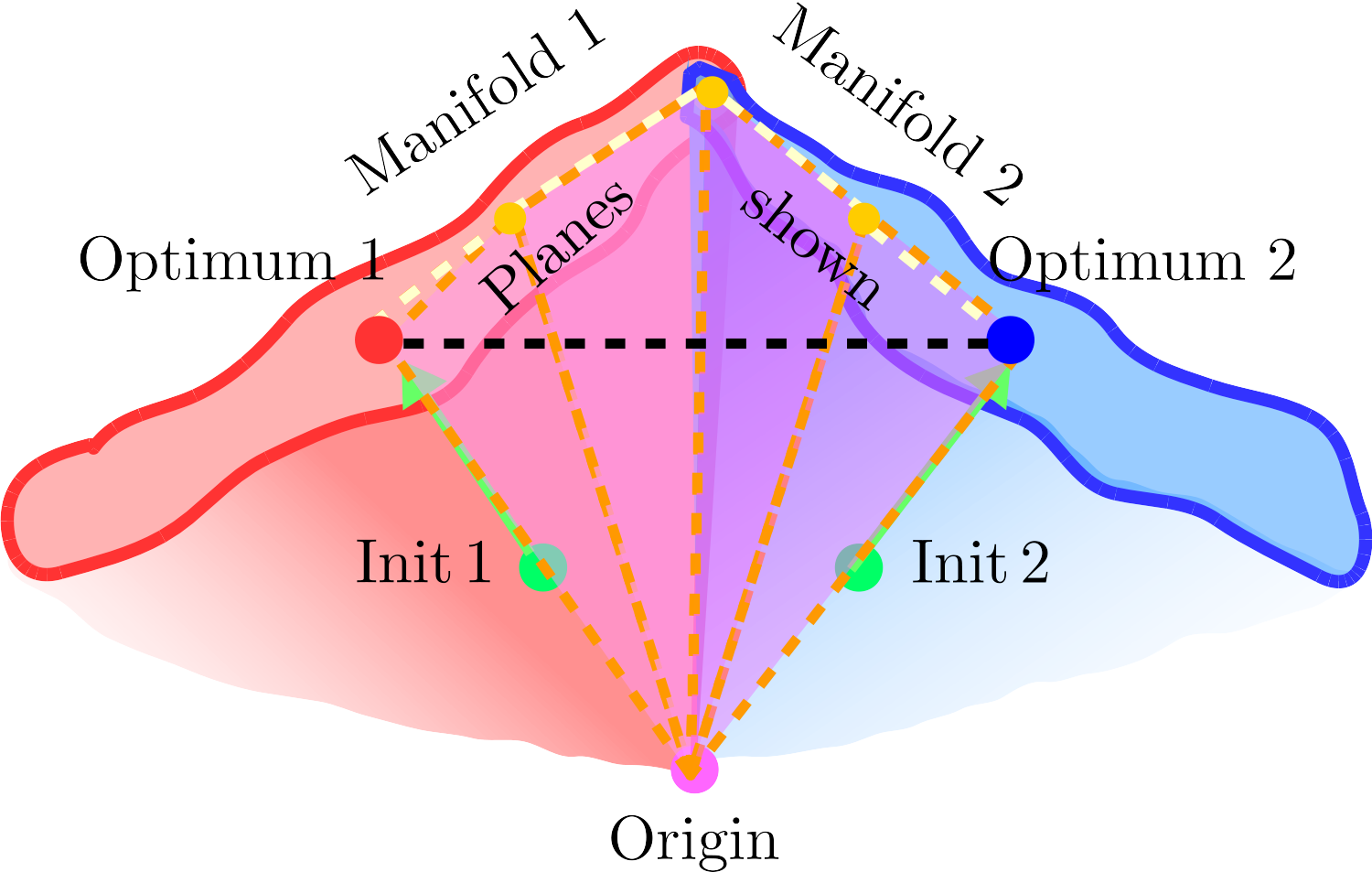}
    }\end{subfigure}
   \begin{subfigure}[Low-loss tunnel]{  
      \includegraphics[width=0.65\textwidth]{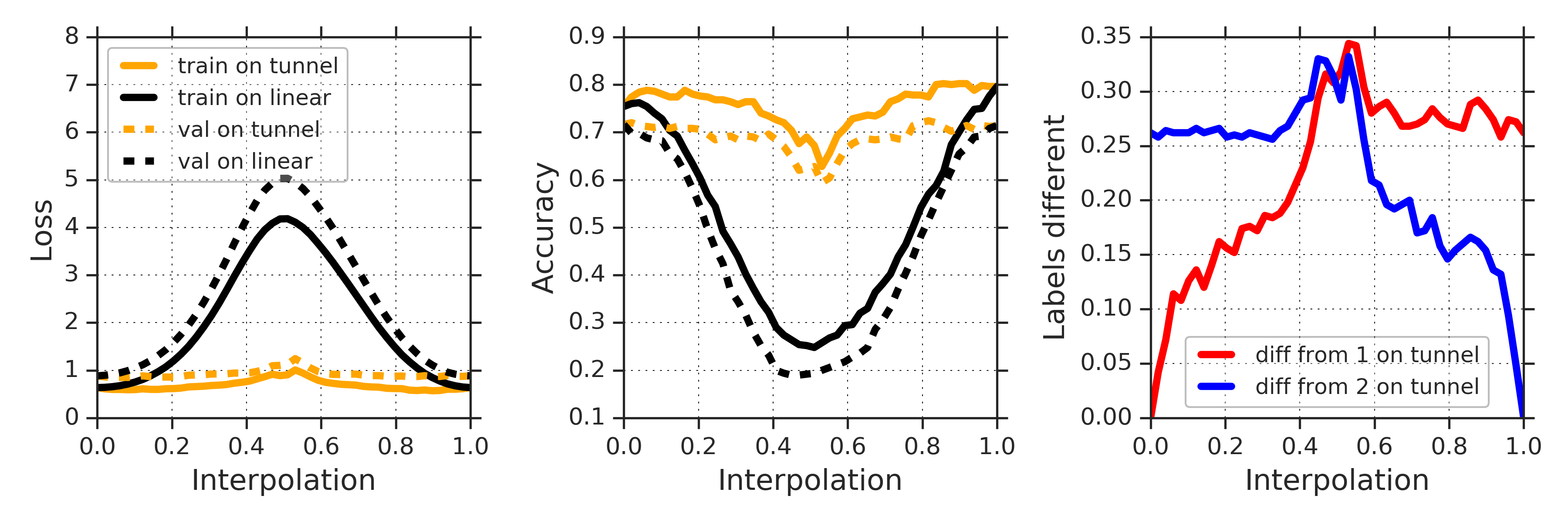}
    }\end{subfigure}
       \reducespaceafterfigure
      \caption{\emph{Left}: Cartoon illustration showing linear connector (black) along with the optimized connector which lies on the manifold of low loss solutions. \emph{Right}: The loss and accuracy in between two independent optima on a linear path and an optimized path in the weight space.
     }
     \label{fig:cartoon:losses_and_accs_on_tunnels}
      \end{center}
\end{figure}

Next, we construct a low-loss tunnel between different optima using the procedure proposed by \citet{fort2019large}, which is a simplification of the procedures proposed in \citet{garipov2018loss} and \citet{draxler2018essentially}. As shown in %
Figure~\ref{fig:cartoon:losses_and_accs_on_tunnels}(a), 
we start at the linear interpolation point (denoted by the black line) and reach the closest point on the manifold by minimizing the training loss. The minima of the training loss are denoted by the yellow line in the manifolds.  
Figure~\ref{fig:cartoon:losses_and_accs_on_tunnels}(b) 
confirms that the tunnel is indeed low-loss. %
This also confirms the findings of \citep{garipov2018loss,fort2019large} that while solutions along the tunnel have similar loss, they are dissimilar in function space.  

\begin{figure}[ht]%
    \centering%
       \begin{subfigure}[Accuracy along the low-loss tunnel]{ 
       \includegraphics[width=\sfigwidthlandscape]{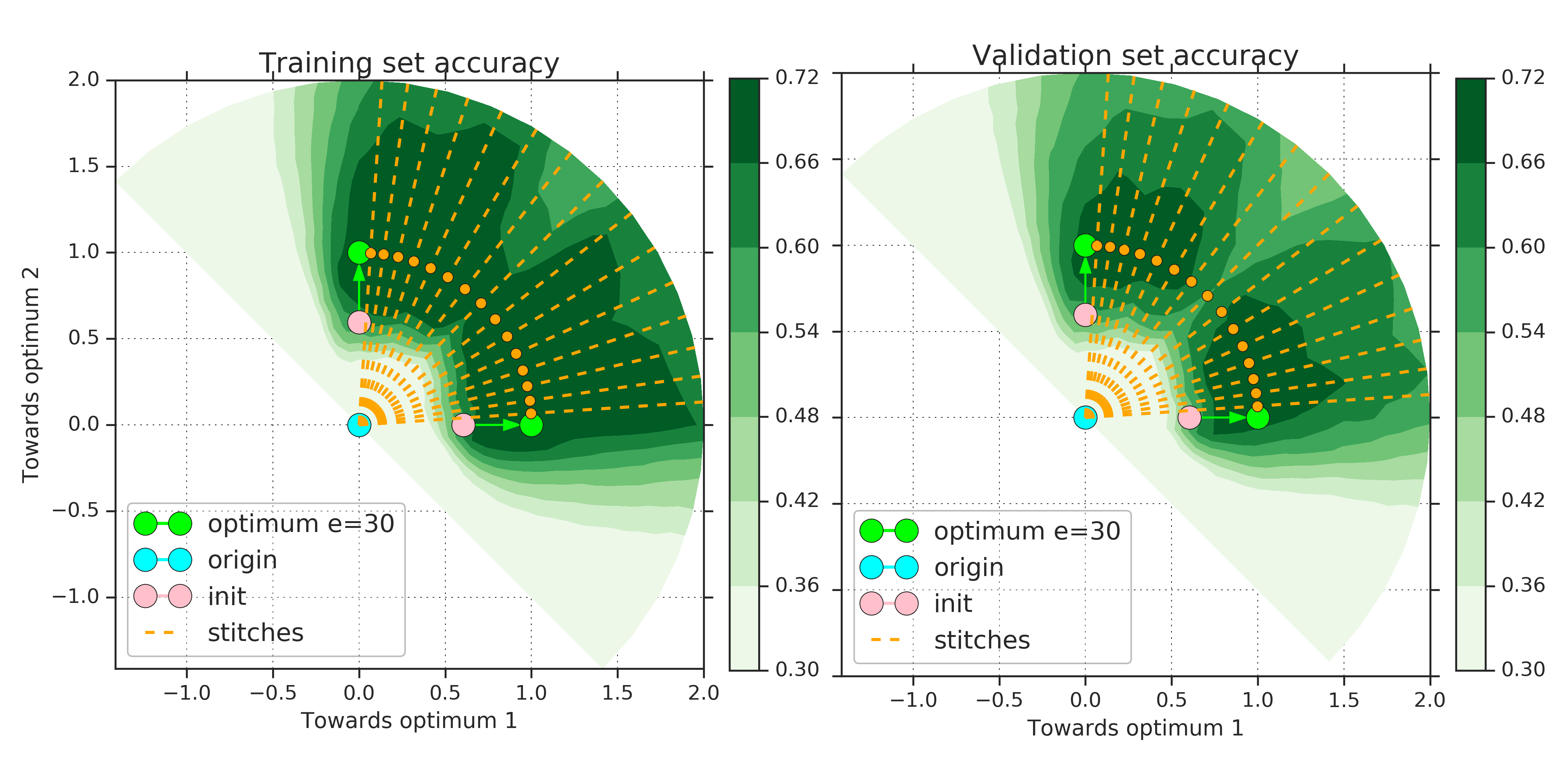}
       }\end{subfigure}
         \begin{subfigure}[Prediction similarity along the low-loss tunnel]{ 
         \includegraphics[width=\sfigwidthlandscape]{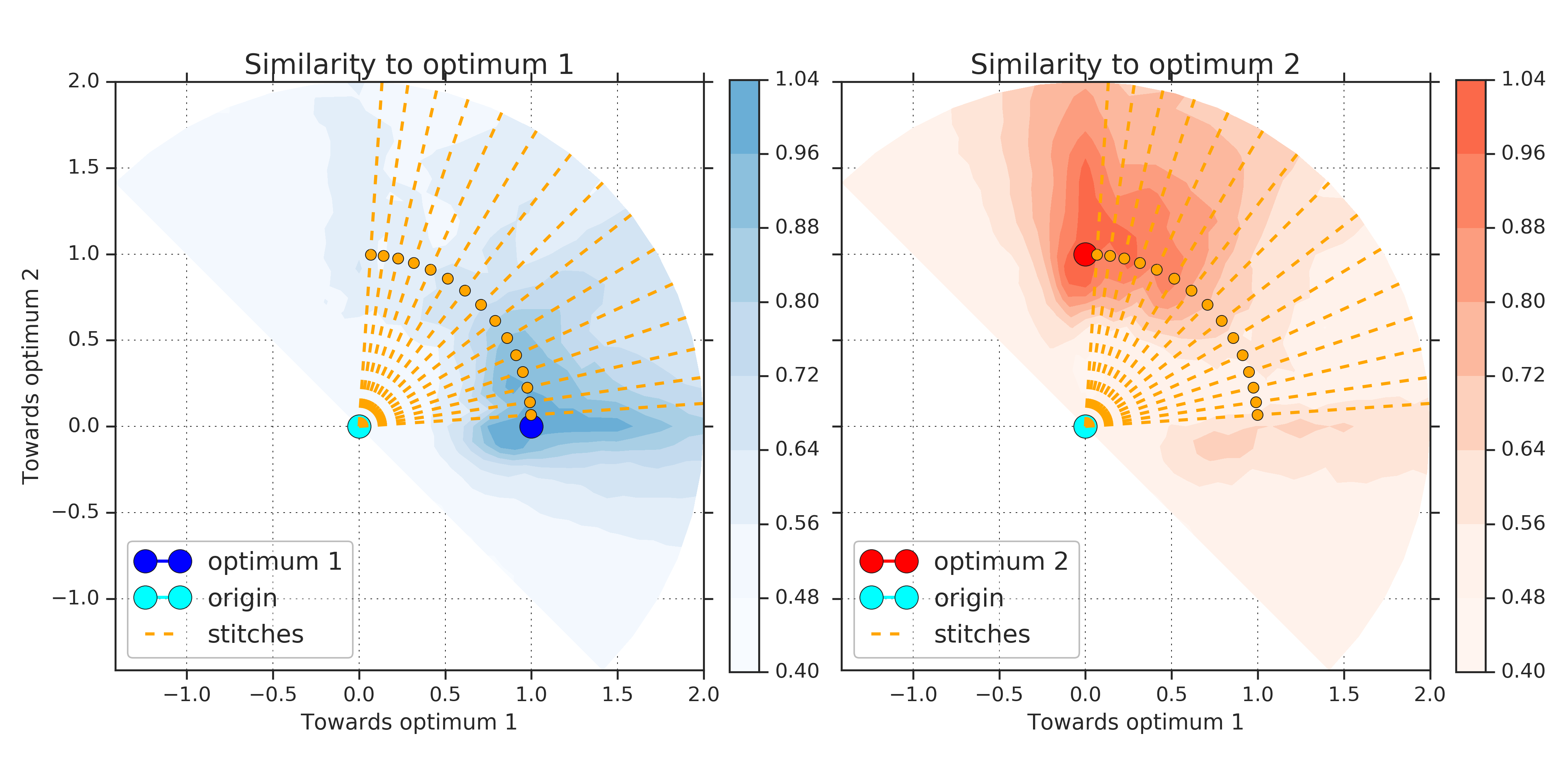}
          }\end{subfigure}
    \caption{\emph{Results using MediumCNN on CIFAR-10}: Radial loss landscape cut between the origin and two independent optima along an optimized low-loss connector and function space similarity (agreement of predictions) to the two optima along the same planes.}
    \label{fig:SimpleCNN_radial_cuts_on_tunnel}
\end{figure}%

In order to visualize the 2-dimensional cut through the loss landscape and the associated predictions along a curved low-loss path, we divide the path into linear segments, and compute the loss and prediction similarities on a triangle given by this segment on one side and the origin of the weight space on the other. We perform this operation on each of the linear segments from which the low-loss path is constructed, and place them next to each other for visualization. Figure~\ref{fig:SimpleCNN_radial_cuts_on_tunnel} visualizes the loss along the manifold, as well as the similarity to the original optima.  Note that the regions between radial yellow lines consist of segments, and we stitch these segments  together in Figure~\ref{fig:SimpleCNN_radial_cuts_on_tunnel}. 
{The accuracy plots show that as we traverse along the low-loss tunnel, the accuracy remains fairly constant as expected. However, the prediction similarity plot shows that the low-loss tunnel does not correspond to similar solutions in function space. 
What it shows is that while the modes are connected in terms of accuracy/loss, their functional forms remain distinct and they do not collapse into a single mode.}

\section{Effect of randomness: random initialization versus random shuffling}\label{sec:randomness}
\begin{figure}[ht]%
    \centering%
      \includegraphics[width=0.7\textwidth]{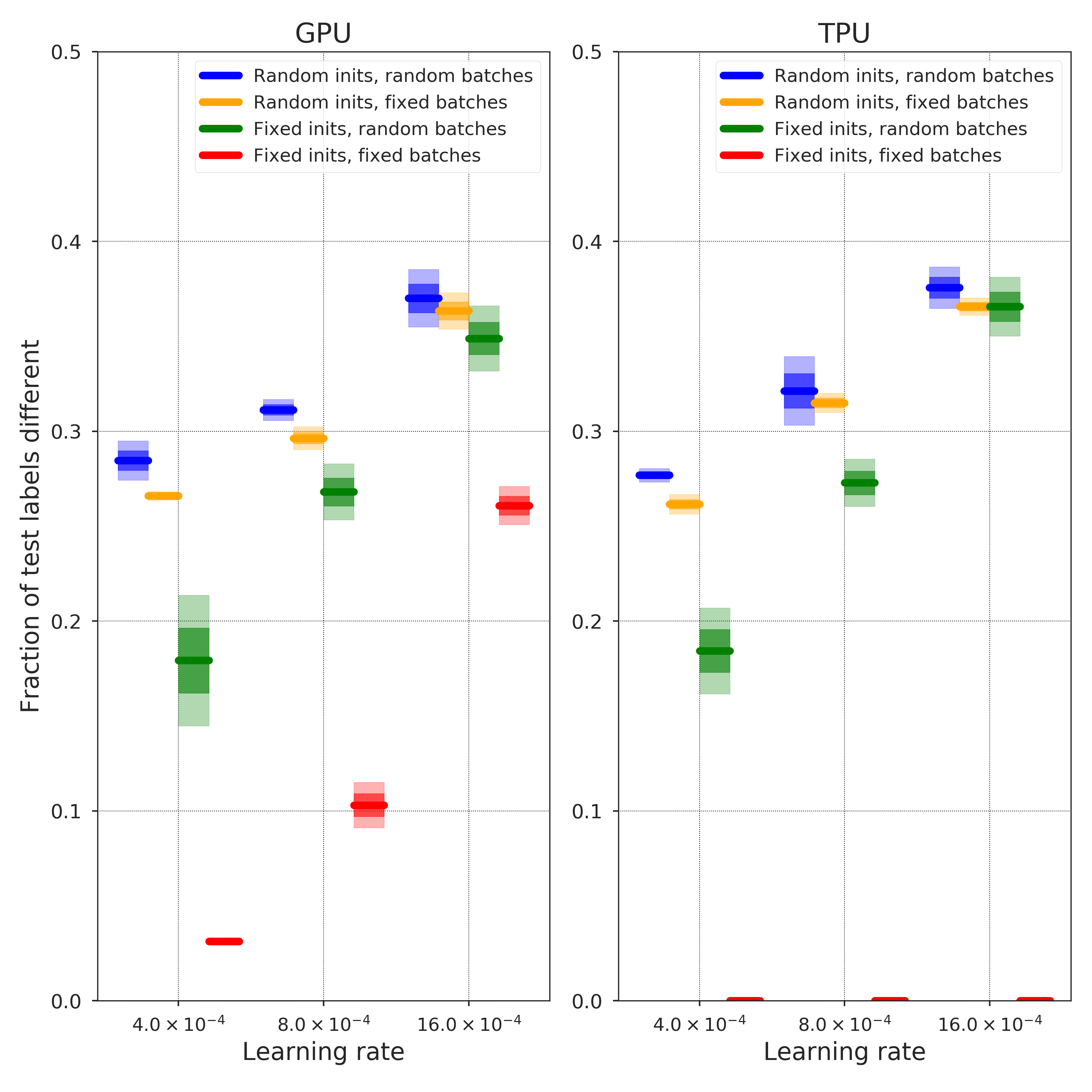}
      \reducespaceafterfigure
    \caption{The effect of random initializations and random training batches on the diversity of predictions. For runs on a GPU, the same initialization and the same training batches (red) do not lead to the exact same predictions. On a TPU, such runs always learn the same function and have therefore 0 diversity of predictions.}
    \label{fig:randomness}%
\end{figure}

Random seed affects both initial parameter values as well the order of shuffling of data points. %
We run experiments to decouple the effect of random initialization and shuffling; Figure~\ref{fig:randomness} shows the results. We observe that both of them provide complementary sources of randomness, with random initialization being the dominant of the two. As expected, random mini-batch shuffling adds more randomness at higher learning rates due to gradient noise. 

\section{Comparison to cSG-MCMC}\label{sec:csgmcmc}
\citet{Zhang2020Cyclical} show that vanilla stochastic gradient Markov Chain Monte Carlo (SGMCMC) methods do not explore multiple modes in the posterior and instead propose cyclic stochastic gradient MCMC (cSG-MCMC) to achieve that. We ran a suite of verification experiments to determine whether the diversity of functions found using the proposed cSG-MCMC algorithm matches that of independently randomly initialized and trained models.  %

\begin{figure*}[ht]
\begin{center}
  \includegraphics[width=0.45\linewidth]{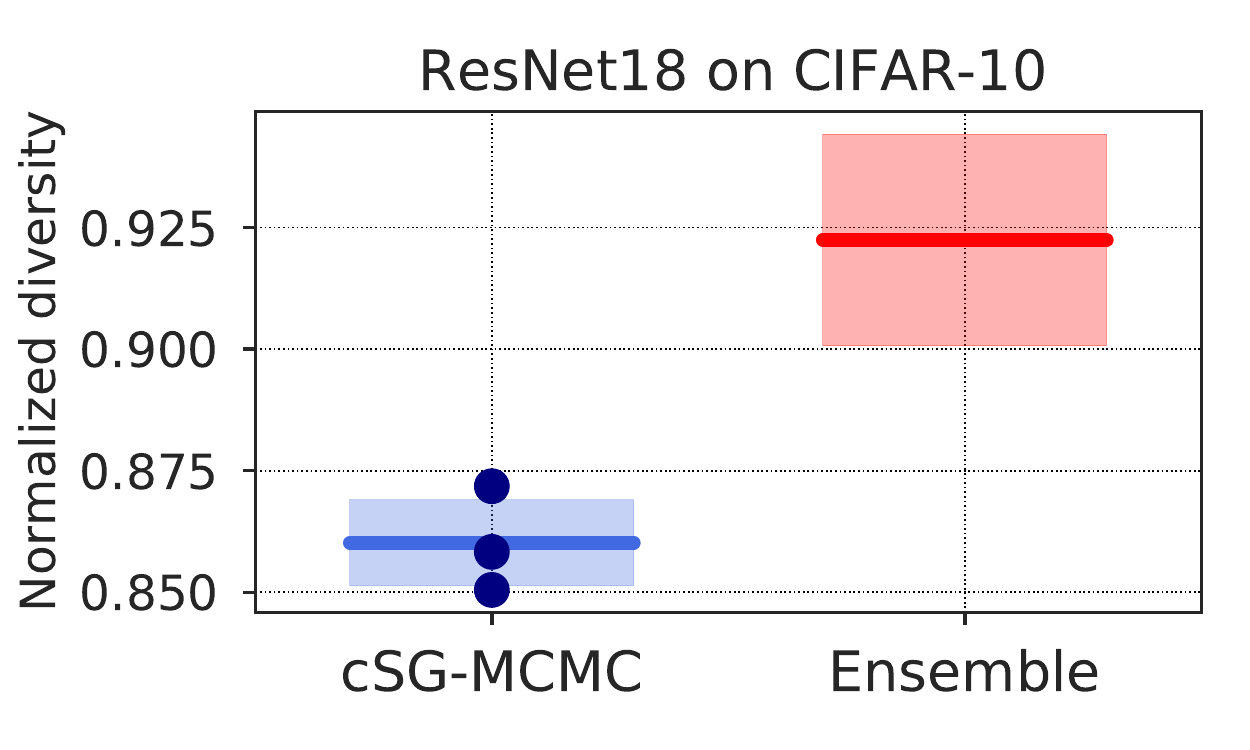}
  \includegraphics[width=0.45\linewidth]{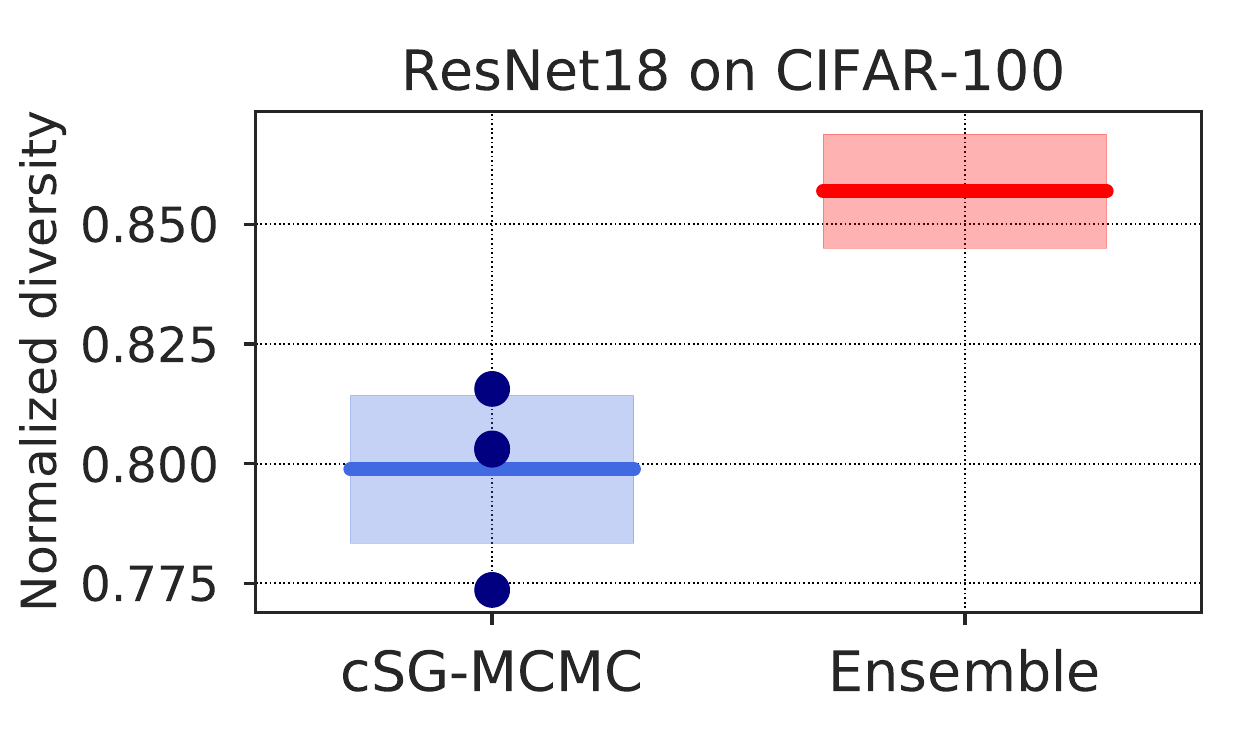}
  \label{fig:cSGMCMC}
  \reducespaceafterfigure
  \caption{Comparison of function space diversities between the cSG-MCMC (blue) and deep ensembles (red). The left panel shows the experiments with ResNet-18 on CIFAR-10 and the right panel shows the experiments on CIFAR-100. In both cases, deep ensembles produced a statistically significantly more diverse set of functions than cSG-MCMC as measured by our function diversity metric. The plots show the mean and 1$\sigma$ confidence intervals based on 4 experiments each.}
  \end{center}
\end{figure*}

We used the code published by the authors \citet{Zhang2020Cyclical} \footnote{ \url{https://github.com/ruqizhang/csgmcmc}} to match exactly the setup of their paper. We ran cSG-MCMC from 3 random initializations, each for a total of 150 epochs amounting to 3 cycles of the 50 epoch period learning rate schedule. We used a ResNet-18 and ran experiments on both CIFAR-10 and CIFAR-100. We measured the function diversity between a) independently initialized and trained runs, and b) between different cyclic learning rate periods within the same run of the cSG-MCMC. The latter (b) should be comparable to the former (a) if cSG-MCMC was as successful as vanilla deep  ensembles at producing diverse functions. We show that both for CIFAR-10 and CIFAR-100, vanilla ensembles generate statistically significantly more diverse sets of functions than cSG-MCMC, as shown in Figure \ref{fig:cSGMCMC}. While cSG-MCMC is doing well in absolute terms, the shared initialization for cSG-MCMC training  seems to lead to lower diversity than deep ensembles with multiple random initializations.  
Another difference between the methods is that individual members of deep ensemble can be trained in parallel unlike cSG-MCMC. 

\reducevspace
\section{Modeling the accuracy -- diversity trade off}
\label{sec:limit:curves}
In our diversity--accuracy plots (e.g. Figure~\ref{fig:diversity_plots_all}), subspace samples trade off their accuracy for diversity in a characteristic way. To better understand where this relationship comes from, we derive  several limiting curves based on an idealized model. We also propose a 1-parameter family of functions that provide a surprisingly good fit (given the simplicity of the model) to our observation, as shown in Figure~\ref{fig:diversity_accuracy_family}. %

We will be studying a pair of functions in a $C$-class problem: the reference solution $f^*$ of accuracy $a^*$, and another function $f$ of accuracy $a$.

\subsection{Uncorrelated predictions: the %
best case}
The best case scenario is when the predicted labels are uncorrelated with the reference solution's labels. On a particular example, the probability that the reference solution got it correctly is $a^*$, and the probability that the new solution got it correctly is $a$. On those examples, the predictions do not differ since they both have to be equal to the ground truth label. The probability that the reference solution is correct on an example while the new solution is wrong is $a^* (1-a)$. The probability that the reference solution is wrong on an example while the new solution is correct is $(1-a^*) a$. 
On the examples where both solutions are wrong (probability $(1-a^*)(1-a)$), there are two cases: 
\vspace{-0.5em}
\begin{enumerate}[itemsep=0em,leftmargin=25pt]
\item the  solutions agree (an additional factor of $1/(C-1)$), or 
\item the two solutions disagree (an additional factor of $(C-2)/(C-1)$).
\end{enumerate}
Only case 2 
contributes to the fraction of labels on which they disagree. Hence we end up with
\begin{align}
d_{\mathrm{diff}}(a; a^*, C) 
= (1-a^*)a + (1-a)a^* + (1-a^*)(1-a) \frac{C-2}{C-1} \, . \nonumber
\end{align}

This curve corresponds to the upper limit in Figure~\ref{fig:diversity_plots_all}. The diversity reached in practice is not as high as the theoretical optimum even for the independently initialized and optimized solutions, which provides scope for future work.  

\subsection{Correlated predictions: the lower limit}
By inspecting Figure~\ref{fig:diversity_plots_all} as well as a priori, we would expect a function $f$ close to the reference function $f^*$ in the weight space to have correlated predictions. We can model this by imagining that the predictions of $f$ are just the predictions of the reference solution $f^*$ perturbed by perturbations of a particular strength (which we vary). 
Let the probability of a label changing be $p$. 
We will consider four cases: %
\begin{enumerate}[itemsep=0em,leftmargin=25pt]
    \item the label of the correctly classified image does not flip (probability $a^* (1-p)$), 
    \item it flips (probability $a^* p$), 
    \item an incorrectly labelled image does not flip (probability $(1-a^*) (1-p)$), and 
    \item it flips (probability $(1-a^*) p$).
\end{enumerate}
The resulting accuracy $a(p)$ obtains a contribution $a^* (1-p)$ from case 1) and a contribution $(1-a^*) p$ with probability $1/(C-1)$ from case 4). Therefore $a(p) = a^* (1-p) + p (1-a^*)/(C-1)$. Inverting this relationship, we get $p(a) = (C-1)(a^* - a)/(Ca^* - 1)$. The fraction of labels on which the solutions disagree is simply $p$ by our definition of $p$, and therefore
\begin{equation}
d_{\mathrm{diff}}(a; a^*, C) = \frac{(C-1)(a^*-a)}{Ca^* - 1} \, .
\end{equation}
This curve corresponds to the lower limit in Figure~\ref{fig:diversity_plots_all}.

\subsection{Correlated predictions: 1-parameter family}
\begin{figure}[ht]%
	\centering%
	\includegraphics[width=1.0\linewidth]{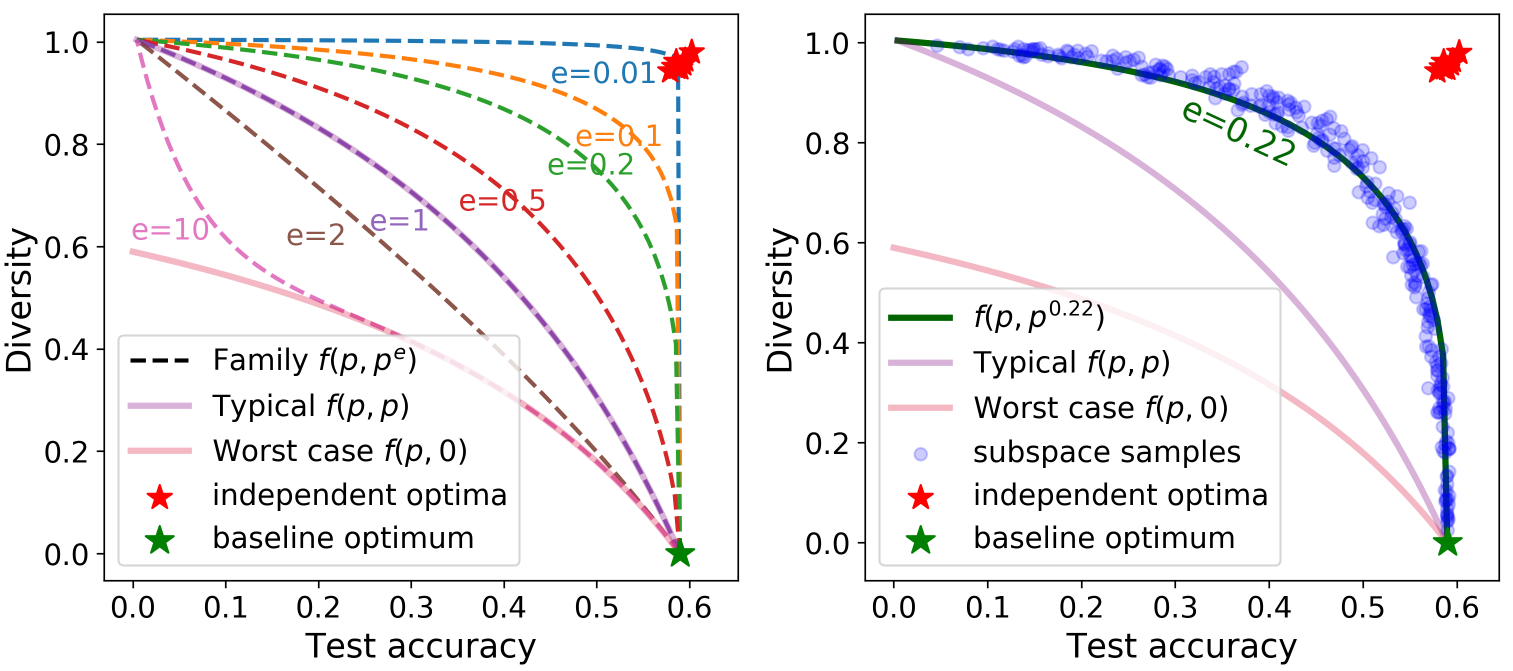}
	\reducespaceafterfigure
	\caption{Theoretical model of the accuracy-diversity trade-off and a comparison to \emph{ResNet20v1} on CIFAR-100. The left panel shows accuracy-diversity trade offs modelled by a 1-parameter family of functions specified by an exponent $e$. The right panel shows real subspace samples for a \emph{ResNet20v1} trained on CIFAR-100 and the best fitting function with $e=0.22$.}
	\label{fig:diversity_accuracy_family}
\end{figure}%
We can improve upon this model by considering two separate probabilities of labels flipping: $p_\mathrm{+}$, which is the probability that a correctly labelled example will flip, and $p_\mathrm{-}$, corresponding to the probability that a wrongly labelled example will flip its label. By repeating the previous analysis, we obtain
 \begin{align}
 &d_{\mathrm{diff}}(p_\mathrm{+},p_\mathrm{-}; a^*, C) %
 = a^* (1-p_\mathrm{+}) + (1-a^*) p_\mathrm{-} \frac{1}{C-1} \, , 
 \end{align}
and
 \begin{equation}
 a(p_\mathrm{+},p_\mathrm{-}; a^*, C) = a^* p_\mathrm{+} + (1-a^*) p_\mathrm{-} \, .
 \end{equation}
 
The previously derived lower limit corresponds to the case $p_\mathrm{+} = p_\mathrm{-} = p \in [0,1]$, where the probability of flipping the correct labels is the same as the probability of flipping the incorrect labels. The absolute worst case scenario would correspond to the situation where $p_\mathrm{+}=p$, while $p_\mathrm{-} = 0$, i.e. only the correctly labelled examples flip.

We found that tying the two probabilities together via an exponent $e$ as $p_\mathrm{+}=p$ and $p_\mathrm{-}=p^e = p_\mathrm{+}^e$ generates a realistic looking trade off between accuracy and diversity. We show the resulting functions for several values of $e$ in Figure~\ref{fig:diversity_accuracy_family}. For $e<1$, the chance of flipping the wrong label is \textit{larger} than that of the correct label, simulating a robustness of the learned solution to perturbations. We found the closest match to our data provided by $e=0.22$.

\end{document}